\documentclass{article}

\usepackage{PRIMEarxiv}

\usepackage[utf8]{inputenc} 
\usepackage[T1]{fontenc}    
\usepackage{hyperref}       
\usepackage{url}            
\usepackage{booktabs}       
\usepackage{amsfonts}       
\usepackage{nicefrac}       
\usepackage{microtype}      
\usepackage{lipsum}
\usepackage{fancyhdr}       
\usepackage{graphicx}       
\graphicspath{{media/}}     
\usepackage[numbers]{natbib}
\usepackage{amsmath}
\usepackage{amssymb}
\usepackage{mathtools}
\usepackage{amsthm}

\usepackage{booktabs}
\usepackage{array}
\usepackage{tabularx}
\usepackage{multicol}
\usepackage{multirow}
\usepackage{makecell}
\usepackage{xcolor}
\usepackage{colortbl}
\usepackage{microtype}

\definecolor{BAShade}{gray}{0.95}
\definecolor{LowShade}{RGB}{255,235,235}
\definecolor{LPTgt}{RGB}{226,122,52}
\definecolor{LPAlt}{RGB}{66,118,186}
\usepackage{tikz}
\usetikzlibrary{decorations.pathreplacing,arrows.meta}

\usepackage{graphicx}
\usepackage{subcaption}

\usepackage{enumitem}
\usepackage{microtype}
\usepackage{url}
\usepackage{tcolorbox}
\usepackage{listings}

\usepackage[normalem]{ulem}

\usepackage{algorithm}
\usepackage{algorithmic}

\usepackage{url}
\usepackage{hyperref}

\usepackage[capitalize,noabbrev]{cleveref}

\definecolor{codegreen}{rgb}{0,0.6,0}
\definecolor{codegray}{rgb}{0.5,0.5,0.5}
\definecolor{codepurple}{rgb}{0.58,0,0.82}
\definecolor{backcolour}{HTML}{F2F5F9}
\definecolor{framecolour}{HTML}{DDE6F0}

\lstdefinestyle{mystyle}{
    backgroundcolor=\color{backcolour},
    commentstyle=\color{codegreen},
    keywordstyle=\color{magenta},
    numberstyle=\tiny\color{codegray},
    stringstyle=\color{codepurple},
    basicstyle=\ttfamily\footnotesize,
    breakatwhitespace=false,
    breaklines=true,
    captionpos=b,
    keepspaces=true,
    numbers=left,
    numbersep=5pt,
    showspaces=false,
    showstringspaces=false,
    showtabs=false,
    tabsize=2,
    escapechar=@
}

\newcommand{\target}[1]{\colorbox{yellow!40}{\textcolor{red}{\textbf{#1}}}}

\newcommand{\squishlist}{
   \begin{list}{$\bullet$}
    { \setlength{\itemsep}{0pt}
      \setlength{\parsep}{0pt}
      \setlength{\topsep}{-3pt}
      \setlength{\partopsep}{0pt}
      \setlength{\listparindent}{-2pt}
      \setlength{\itemindent}{-5pt}
      \setlength{\leftmargin}{1em}
      \setlength{\labelwidth}{0em}
      \setlength{\labelsep}{0.5em} } }

\newcommand{\squishend}{
    \end{list}
}

\title{
LP-SFT: Keeping Plausible Alternatives Alive in Supervised Fine-Tuning
}

\author{
Yueyang Wang$^{1}$\thanks{Corresponding author: \texttt{wangyueyang@stu.pku.edu.cn}}
\quad
Baolong Bi$^{2}$
\quad
Shuo Lu$^{3}$
\quad
Jingyuan Zhang$^{4}$
\quad
Jiajun Shi$^{5}$
\quad
Yuxuan Zhang$^{6,7}$
\\[0.5em]
$^{1}$School of Mathematical Sciences, Peking University
\\
$^{2}$Institute of Computing Technology, Chinese Academy of Sciences
\\
$^{3}$Institute of Automation, Chinese Academy of Sciences
\\
$^{4}$College of Computing, Georgia Institute of Technology
\\
$^{5}$CCSE, Beihang University
\\
$^{6}$University of British Columbia, $^{7}$Vector Institute
}

\begin{document}
\maketitle

\begin{abstract} 
\looseness=-1 
Supervised fine-tuning (SFT) is widely used to adapt pretrained language models to downstream domains, but can over-specialize the model and degrade its pre-existing capabilities. Standard cross-entropy drives probability mass onto the observed target token and can make plausible alternatives that the pretrained model itself endorsed vanish. Using Shannon and R\'{e}nyi entropies, we show that pretrained next-token distributions exhibit a regular multimodal structure, with entropy peaks corresponding to different numbers of plausible alternatives. Motivated by this observation, we propose \textbf{LP-SFT}, a \textbf{L}ocal-\textbf{P}reserving \textbf{S}upervised \textbf{F}ine-\textbf{T}uning objective with two key design principles: it removes the supervised token from a local top-$K$ candidate set to avoid conflict with cross-entropy, and applies a locally normalized KL loss to preserve relative preferences among the remaining non-label alternatives. Across mixed-domain and domain-specific fine-tuning experiments, LP-SFT consistently outperforms standard SFT and recent baselines in aggregate performance while keeping base-endorsed alternatives from vanishing, achieving a favorable balance between single-sample accuracy and finite-budget solution accessibility, as measured by pass@1 and pass@$k$, respectively, with only a modest increase in training cost.
\end{abstract}


\section{Introduction}

Supervised fine-tuning (SFT) has become the dominant paradigm for adapting pretrained language models to instruction following, reasoning, coding, and domain-specific tasks~\citep{ouyang2022training, wang2023self}. Despite its simplicity and effectiveness, SFT can degrade pre-existing capabilities in two ways: over-specialization to the fine-tuning distribution can cause catastrophic forgetting of out-of-domain capabilities, while gains in single-sample accuracy can make correct solutions less accessible on challenging problems~\citep{welleck2019neural, luo2025empirical}.\looseness=-1

We argue that a significant portion of this degradation stems from the mismatch between the standard cross-entropy objective and the rich distributional knowledge encoded in pretrained models. Cross-entropy supervises each position with a single observed token and drives probability mass onto it, even when the pretrained next-token distribution holds several equally plausible continuations. Its effect is therefore not only to promote the target, but to abandon alternatives the model itself endorsed. We take this abandonment to be a principal source of forgetting: what SFT destroys is the set of continuations the pretrained model still considered reasonable. Using Shannon and R\'{e}nyi entropies, we observe a multimodal structure with peaks near $\ln k$, corresponding to near-uniform plateau states over $k$ plausible candidates. This structure persists across model families and datasets, showing that such alternatives are widespread rather than incidental, and that the knowledge at stake lies in the relative preferences among non-label candidates rather than in the target alone.

\begin{figure}[t]
    \centering
    \resizebox{\linewidth}{!}{\begin{tikzpicture}[x=1cm,y=1cm,font=\footnotesize]
\tikzset{
  tgt/.style={fill=LPTgt,draw=LPTgt!75!black,line width=0.3pt},
  alt/.style={fill=LPAlt,draw=LPAlt!75!black,line width=0.3pt},
  tl/.style={fill=gray!40,draw=gray!65,line width=0.3pt},
  ghost/.style={draw=black!40,densely dashed,line width=0.35pt},
  card/.style={draw=black!22,fill=white,rounded corners=3pt,line width=0.5pt},
}
\def\H{2.2}   
\def\h{0.95}  
\def\bw{0.40}
\def\yz{0.42} 
\newcommand{\good}{{\color{green!45!black}\ensuremath{\checkmark}}}
\newcommand{\bad}{{\color{red!65!black}\ensuremath{\times}}}
\def\lpcard#1#2#3#4#5#6{%
\begin{scope}[xshift=#1cm]
  \draw[card] (0,-0.36) rectangle (4.3,2.45);
  \begin{scope}
    \clip[rounded corners=3pt] (0,-0.36) rectangle (4.3,2.45);
    \fill[#3] (0,2.12) rectangle (4.3,2.45);
    \fill[black!3] (0,-0.36) rectangle (4.3,0.0);
  \end{scope}
  \draw[black!22,line width=0.5pt] (0,2.12) -- (4.3,2.12);
  \draw[black!15,line width=0.4pt] (0,0.0) -- (4.3,0.0);
  \node[font=\small] at (2.15,2.285) {#2};
  \node[font=\scriptsize] at (2.15,-0.18) {#6};
  \fill[LPAlt!9,rounded corners=1.5pt] (0.74,\yz) rectangle (3.04,1.1);
  \node[anchor=north east,font=\scriptsize,LPAlt!80!black] at (3.04,1.1) {$\mathcal{A}_t$};
  \ifnum#5=1
    \foreach \v/\x in {0.25/0.86,0.24/1.42,0.23/1.98}
      \draw[ghost] (\x,\yz) rectangle ++(\bw,\v*\H);
  \fi
  \fill[tgt] (0.30,\yz) rectangle ++(\bw,\Tg*\H);
  \fill[alt] (0.86,\yz) rectangle ++(\bw,\Va*\H);
  \fill[alt] (1.42,\yz) rectangle ++(\bw,\Vb*\H);
  \fill[alt] (1.98,\yz) rectangle ++(\bw,\Vc*\H);
  \fill[tl]  (2.54,\yz) rectangle ++(\bw,\Vt*\H);
  \draw[black!70,line width=0.45pt] (0.14,\yz) -- (3.12,\yz);
  \foreach \lab/\x in {$y_t$/0.30,$v_1$/0.86,$v_2$/1.42,$v_3$/1.98,$\cdots$/2.54}
    \node[anchor=north,inner sep=1.5pt] at (\x+0.5*\bw,\yz) {\lab};
  \draw[fill=LPAlt!9,draw=LPAlt!45,rounded corners=2pt,line width=0.4pt] (1.40,1.28) rectangle (4.16,2.04);
  \draw[->,LPAlt!60,line width=0.5pt] (1.90,1.12) -- (1.90,1.26);
  \fill[alt] (1.55,1.36) rectangle ++(0.28,\Ha*\h);
  \fill[alt] (1.92,1.36) rectangle ++(0.28,\Hb*\h);
  \fill[alt] (2.29,1.36) rectangle ++(0.28,\Hc*\h);
  \fill[tl]  (2.66,1.36) rectangle ++(0.28,\Ht*\h);
  \draw[black!60,line width=0.35pt] (1.48,1.36) -- (3.02,1.36);
  \node[anchor=west,font=\scriptsize,inner sep=0pt] at (3.08,1.66) {#4};
\end{scope}}
\def\Tg{0.26}\def\Va{0.25}\def\Vb{0.24}\def\Vc{0.23}\def\Vt{0.02}
\def\Ha{0.34}\def\Hb{0.32}\def\Hc{0.31}\def\Ht{0.03}
\lpcard{0}{Base model $q_t$}{black!6}{$\hat q_t$}{0}{plateau-4: four near-equal continuations}
\def\Tg{0.62}\def\Va{0.23}\def\Vb{0.09}\def\Vc{0}\def\Vt{0}
\def\Ha{0.66}\def\Hb{0.26}\def\Hc{0}\def\Ht{0}
\lpcard{4.8}{Cross-entropy}{LPTgt!14}{{\color{red!65!black}$\hat p_t\neq\hat q_t$}}{1}{pass@1 \good\qquad pass@$k$ \bad}
\node[font=\small,inner sep=0pt] at (4.8+1.98+0.5*\bw,\yz+0.27) {\bad};
\def\Tg{0.62}\def\Va{0.128}\def\Vb{0.123}\def\Vc{0.118}\def\Vt{0.010}
\def\Ha{0.34}\def\Hb{0.32}\def\Hc{0.31}\def\Ht{0.03}
\lpcard{9.6}{LP-SFT (ours)}{LPAlt!16}{{\color{green!45!black}$\hat p_t=\hat q_t$}}{1}{pass@1 \good\qquad pass@$k$ \good}
\node[anchor=west,font=\scriptsize,text=black!65] at (9.6+3.08,0.72) {$P_t$ free};
\draw[-{Stealth[length=5pt]},black!55,line width=0.9pt] (4.36,1.1) -- (4.74,1.1);
\draw[black!45,line width=0.4pt] (4.8,2.58) -- (4.8,2.66) -- (13.9,2.66) -- (13.9,2.58);
\node[fill=white,inner sep=2pt,font=\scriptsize,text=black!70] at (9.35,2.66) {after fine-tuning on $y_t$};
\end{tikzpicture}}
    \caption{
    \textbf{Illustration of LP-SFT at a plateau position.}
    $\mathcal{A}_t$: top-$K$ base candidates excluding $y_t$; insets: normalized profile on $\mathcal{A}_t$; dashed: base heights. LP-SFT constrains only the shape of the alternatives and leaves their mass $P_t$ free, so $y_t$ gains as much as under cross-entropy (pass@1) while no alternative vanishes (pass@$k$).
    }
    \label{fig:lp_sft_overview}
    \vspace{-4mm}
\end{figure}

Motivated by this structure, we propose \textbf{LP-SFT} (\textbf{L}ocal-\textbf{P}reserving \textbf{S}upervised \textbf{F}ine-\textbf{T}uning). We first select a local top-$K$ candidate set from the base distribution to capture plausible next-token alternatives at each position, with $K$ motivated by the observed entropy modes. LP-SFT then follows two design principles. First, it removes the supervised token from this candidate set so that the preservation objective does not oppose cross-entropy. Second, it locally normalizes the base and trainable distributions over the resulting non-label set, preserving relative preferences without constraining its total probability mass, which prevents plausible candidates from vanishing (Figure~\ref{fig:lp_sft_overview}). Gradient analysis and ablations validate these design choices, while the fixed top-$K$ reference can be cached offline, avoiding the need for a frozen reference model during training.\looseness=-1

We evaluate LP-SFT on three base models and three fine-tuning datasets covering general instruction following, code, and mathematics. LP-SFT consistently improves aggregate performance over standard SFT and recent SFT objectives and achieves the best overall pass@$k$ performance. An additional Qwen2.5-72B scaling study confirms that these gains persist at substantially larger scale: LP-SFT has fine-tuning cost comparable to standard SFT, plus a one-time, reusable cache-construction cost.\looseness=-1

Our contributions are summarized as follows:
\squishlist
\item We provide a systematic analysis of pretrained next-token distributions, revealing multimodal entropy structure and heterogeneous uncertainty regimes across token positions.

\item We propose \textbf{LP-SFT}, an efficient objective with two complementary design principles: target-token removal avoids conflict with cross-entropy, while local normalization preserves relative preferences among plausible non-label alternatives.

\item Extensive experiments across model families and scales up to 72B show that LP-SFT consistently improves aggregate performance over standard SFT and recent baselines, achieving a favorable balance among adaptation, capability retention, and finite-budget solution accessibility.
\squishend

\section{Related Work}

\paragraph{Improved SFT objectives.}
Recent work has sought to improve SFT beyond standard cross-entropy. Dynamic Fine-Tuning (DFT)~\citep{wu2025generalization} and Entropy-Adaptive Fine-Tuning (EAFT)~\citep{diao2026entropy} modulate the target-token learning signal based on predicted probability or entropy. GEM~\citep{li2025preserving} reduces overfitting via entropy-regularized distribution matching. Most closely related is Anchored Supervised Fine-Tuning (ASFT)~\citep{zhu2025anchored}, which anchors the full predictive distribution to the base model via a KL penalty; this also constrains the target's mass and requires an online frozen reference model. Decoupled Knowledge Distillation~\citep{zhao2022decoupled} likewise separates target and non-target terms, but for teacher--student transfer. LP-SFT instead removes the target from a cached top-$K$ base set and matches only the normalized profile of the remaining alternatives, leaving target supervision unaffected.

\paragraph{Entropy structure in next-token prediction.}
Recent work~\citep{wang2026he} observes that the Shannon entropy of next-token distributions exhibits a multimodal structure, with peaks near $\ln k$ for small integers $k=1,2,3$. Since a uniform distribution over $k$ options has entropy $\ln k$, these peaks indicate states in which uncertainty is spread over $k$ similarly plausible next-token candidates, known as \emph{entropy-compressed states}. Related work also shows that token-level entropy is an informative training signal: high-entropy tokens often correspond to branching points in reasoning, where multiple continuations are plausible and small changes can substantially affect the final outcome~\citep{wang2026beyond}.\looseness=-1

\section{Preliminaries}
\label{sec:preliminaries}

Given an input prompt $\mathbf{x}$ and a target response sequence
$\mathbf{y}=(y_1,\dots,y_T)$, the model estimates the conditional probability of the next response token $y_t$ given the prompt and the preceding response tokens $\mathbf{y}_{<t}=(y_1,\dots,y_{t-1})$. Here, $y_t$ denotes the ground-truth response token at position $t$. Let $\mathcal{V}$ denote the vocabulary. At each position $t$, the model outputs a next-token probability distribution
$p_t(\cdot)=p_\theta(\cdot \mid \mathbf{x}, \mathbf{y}_{<t})$ over $\mathcal{V}$.

\textbf{Top-$k$ Candidate Set.}
We define the top-$k$ candidate set as
\begin{equation}
\label{eq:candidate_set}
    C_k(p_t)=\{v_t^{(1)}, v_t^{(2)}, \dots, v_t^{(k)}\},
\end{equation}
where $v_t^{(i)}$ denotes the token with the $i$-th highest probability under $p_t$ at position $t$. These tokens are sorted in descending order: $p_t(v_t^{(1)}) \ge p_t(v_t^{(2)}) \ge \dots \ge p_t(v_t^{(k)})$.

\textbf{Top-$K$ Normalized Distribution.}
For entropy analysis, we use a fixed top-$K$ truncation window and normalize the probability mass within $C_K(p_t)$. Unless otherwise specified, we set $K=30$. The normalized probability of the $i$-th ranked token is $\hat{p}_{t,i} = p_t(v_t^{(i)}) / \sum_{j=1}^{K} p_t(v_t^{(j)})$ for $i\in\{1,\dots,K\}$.

\textbf{Top-$K$ Shannon Entropy.}
We define the Shannon entropy $H_1$ and its effective support size $N_1$ over the normalized top-$K$ distribution~\citep{hill1973diversity,jost2006entropy}:
\begin{equation}
\label{eq:shannon_effective_support}
H_1(p_t)
=
-\sum_{i=1}^{K} \hat{p}_{t,i}\ln \hat{p}_{t,i},
\qquad
N_1(p_t)
=
\exp(H_1(p_t)).
\end{equation}
The effective support $N_1$ is the equivalent number of uniformly likely candidates, while Shannon entropy measures probability spread within the normalized top-$K$ set.

\textbf{Top-$K$ R\'enyi-2 Entropy.}
We also use the R\'{e}nyi-2 entropy, or collision entropy, to characterize concentration among high-probability candidates. Its effective support size $N_2$ is defined as
\begin{equation}
\label{eq:renyi2_effective_support}
H_2(p_t)
=
-\ln \sum_{i=1}^{K} \hat{p}_{t,i}^{\,2},
\qquad
N_2(p_t)
=
\exp(H_2(p_t))
=
\frac{1}{\sum_{i=1}^{K}\hat{p}_{t,i}^{\,2}}.
\end{equation}
\newlength{\densityH}
\setlength{\densityH}{0.366\textwidth}
\begin{figure*}[!t]
    \centering
    \setlength{\tabcolsep}{2pt}

    \begin{subfigure}[t]{0.305\textwidth}
        \centering
        \includegraphics[height=\densityH]{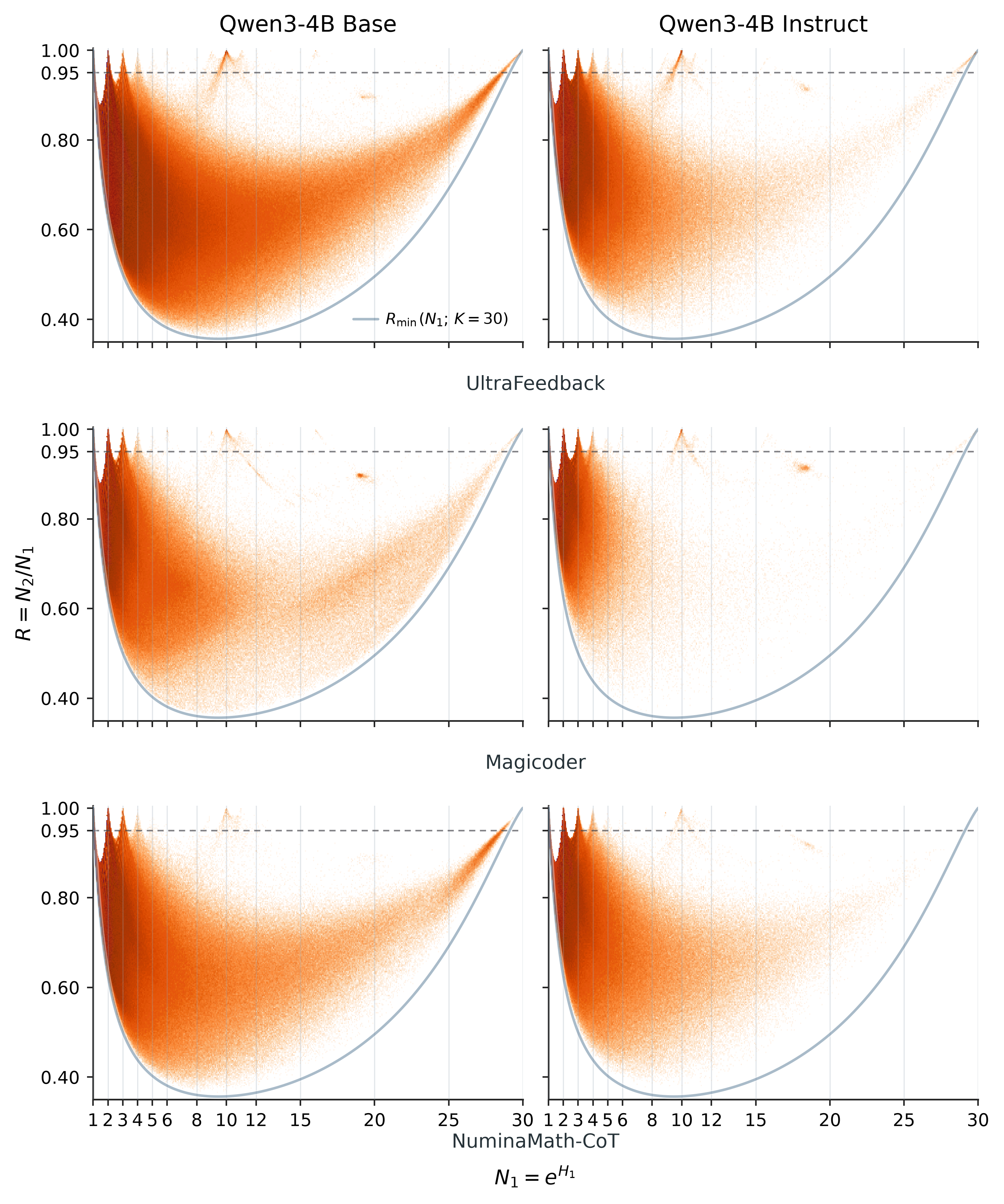}
        \caption{Qwen3-4B}
        \label{fig:qwen3_4b_density}
    \end{subfigure}
    \hfill
    \begin{subfigure}[t]{0.305\textwidth}
        \centering
        \includegraphics[height=\densityH]{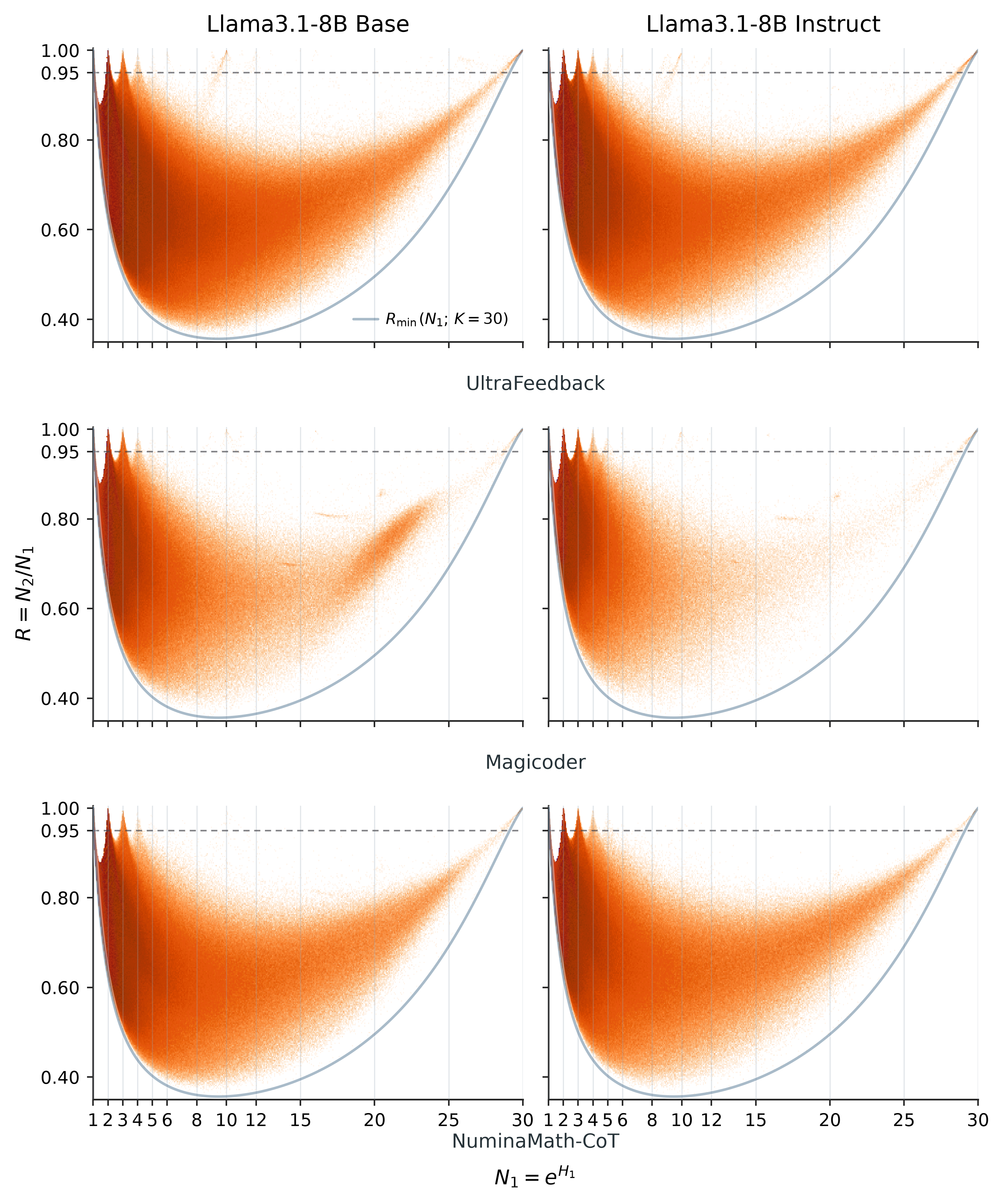}
        \caption{Llama-3.1-8B}
        \label{fig:llama3_1_8b_density}
    \end{subfigure}
    \hfill
    \begin{subfigure}[t]{0.34\textwidth}
        \centering
        \includegraphics[height=\densityH]{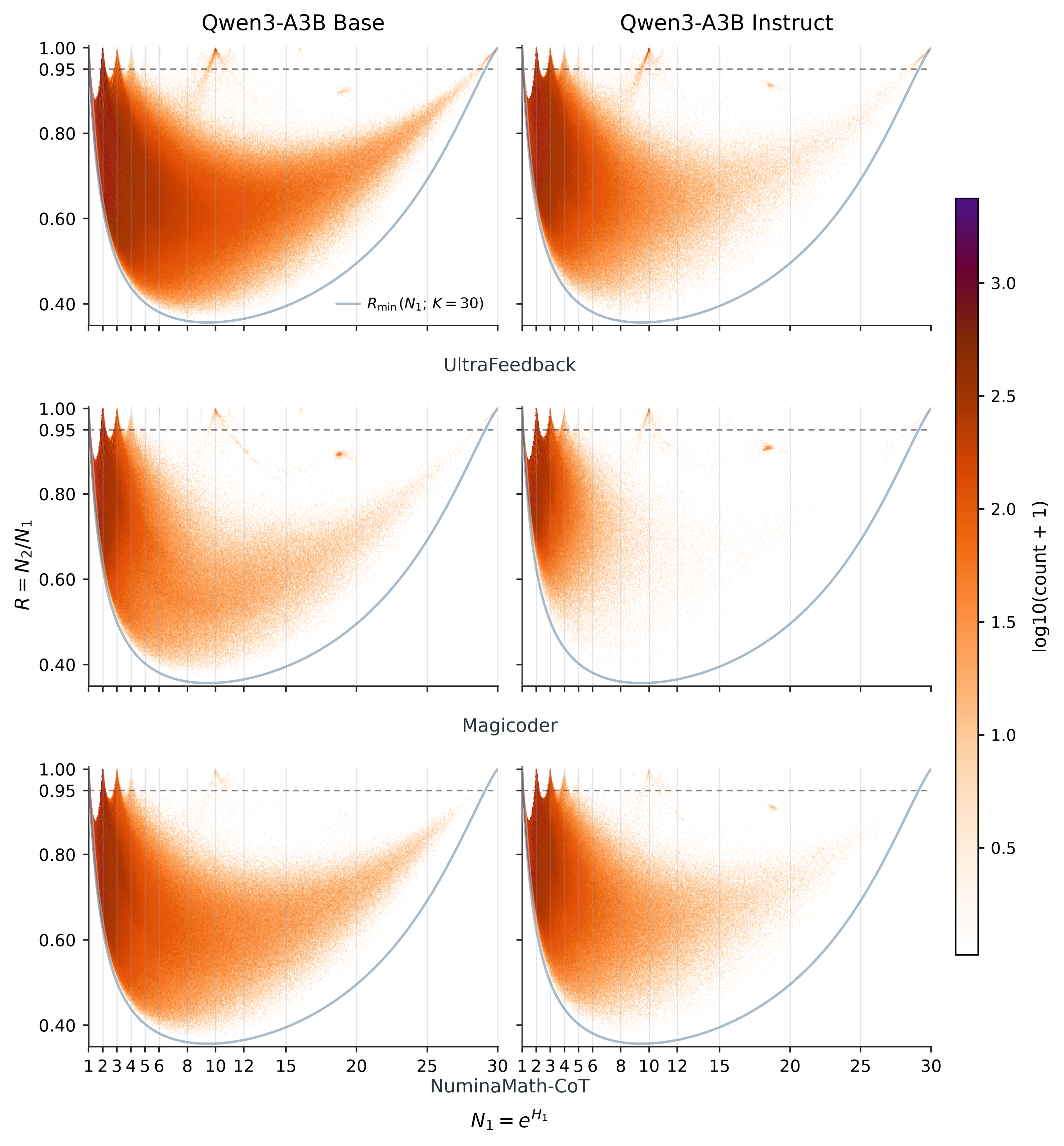}
        \caption{Qwen3-30B-A3B}
        \label{fig:qwen3_a3b_density}
    \end{subfigure}

    \caption{Joint distributions of $N_1$ and $R=N_2/N_1$. High-$R$ ridges at multiple integer support sizes reveal higher-order entropy peaks across datasets and model architectures.
}
    \label{fig:plateau}
\end{figure*}

Compared with Shannon entropy, R\'{e}nyi-2 entropy places more emphasis on high-probability tokens and is therefore more sensitive to concentration around dominant candidates. Throughout the paper, we use $N_1$ and $N_2$ to summarize the effective number of plausible next-token candidates under the normalized top-$K$ distribution.

\section{Multimodal Entropy Structure in Next-Token Distributions}
\label{sec:entropy_structure}

\begin{figure*}[!t]
    \centering

    \begin{subfigure}[t]{0.32\textwidth}
        \centering
        \includegraphics[width=\linewidth]{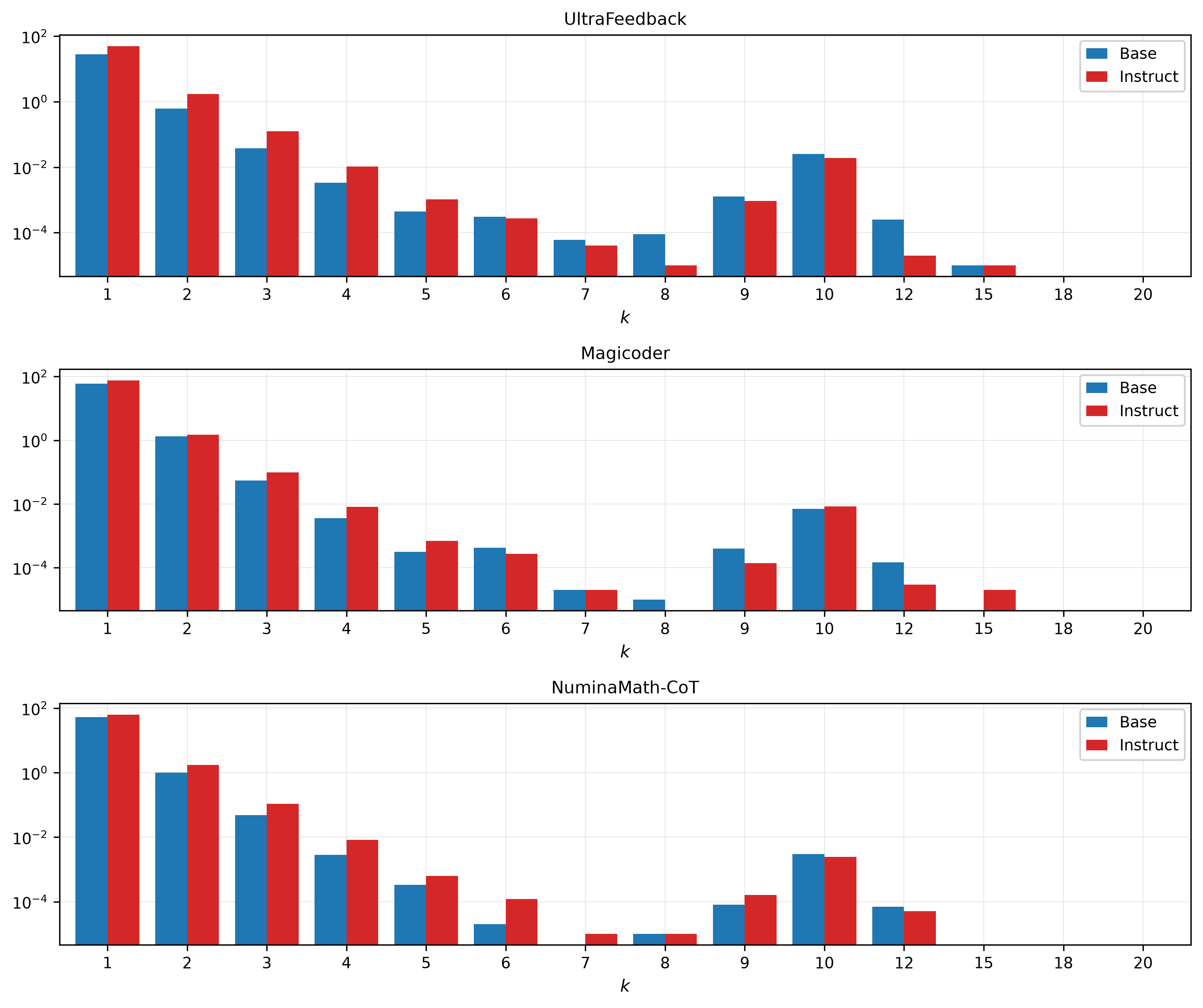}
        \caption{Qwen3-4B}
        \label{fig:qwen3_4b_plateau_k}
    \end{subfigure}
    \begin{subfigure}[t]{0.32\textwidth}
        \centering
        \includegraphics[width=\linewidth]{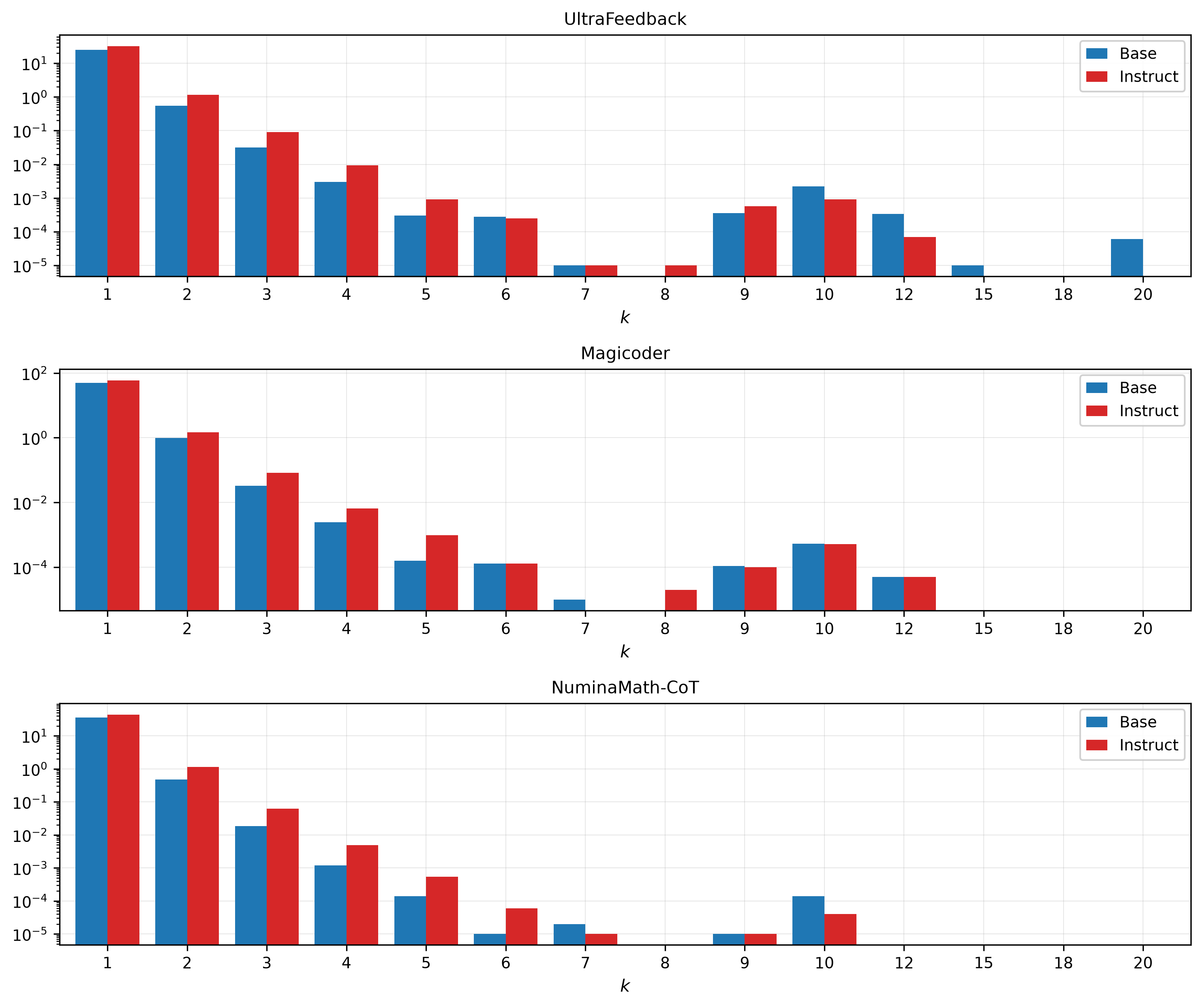}
        \caption{Llama-3.1-8B}
        \label{fig:llama3_1_8b_plateau_k}
    \end{subfigure}
    \begin{subfigure}[t]{0.32\textwidth}
        \centering
        \includegraphics[width=\linewidth]{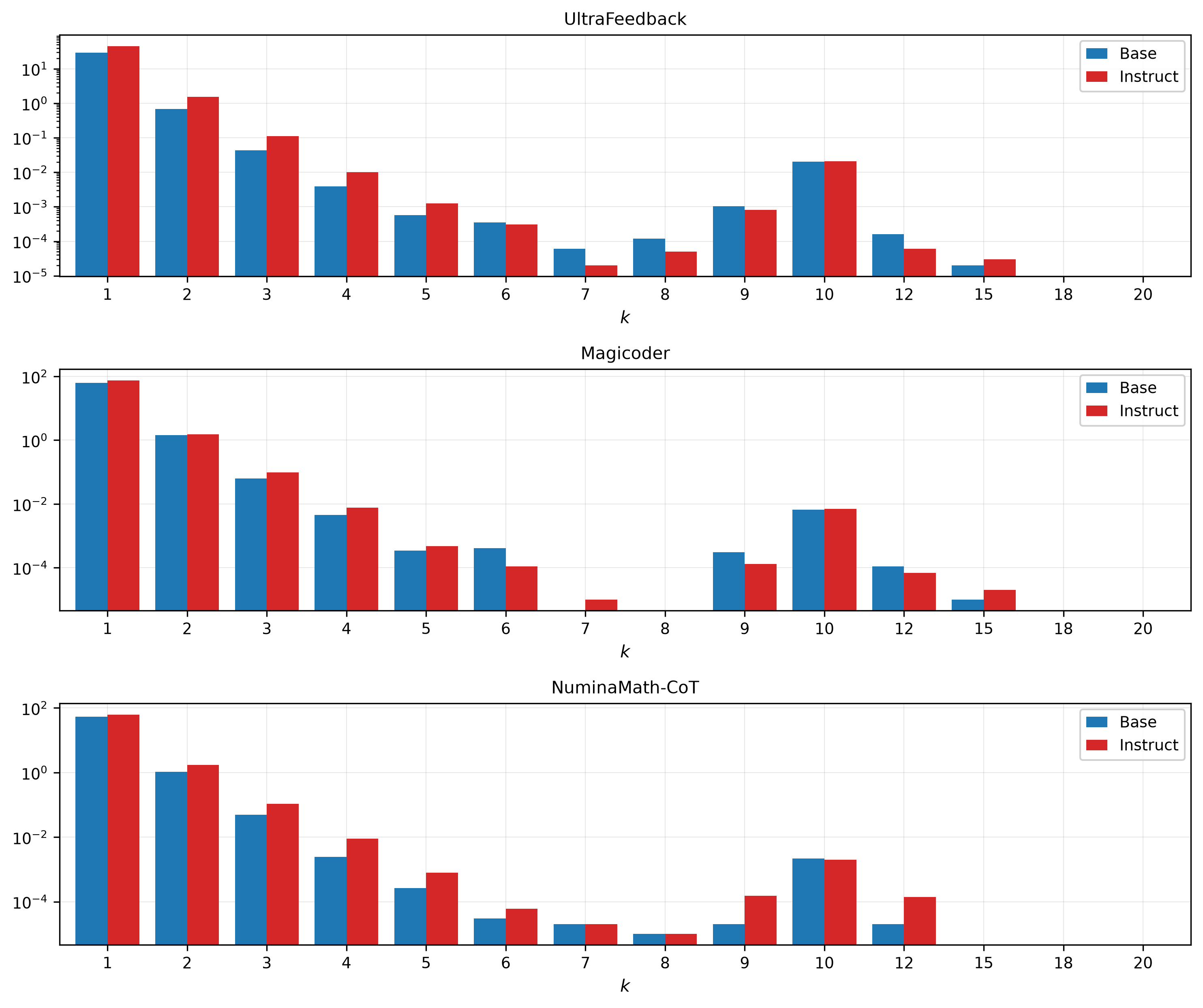}
        \caption{Qwen3-30B-A3B}
        \label{fig:qwen3_30b_a3b_plateau_k}
    \end{subfigure}

    \caption{
    Occurrence of plateau-$k$ states, defined by
    $N_1 \in [k-0.3,k+0.3]$ and $R>0.95$.
    Instruction tuning generally increases low-order plateaus ($k=1$--$5$), while the trend weakens or reverses at higher orders ($k>5$).
    }
    \label{fig:plateau_k}
\end{figure*}

\subsection{Effective Support Ratio for Identifying Entropy Peaks}

While prior work has identified entropy-compressed states as evidence of multimodal structure in next-token prediction, existing analyses mainly characterize low-order peaks up to $\ln 3$, leaving higher-order entropy-compressed states insufficiently studied. To systematically identify such peaks, we introduce the \emph{effective support ratio}:
\begin{equation}
    R(p_t)
    =
    \frac{N_2(p_t)}{N_1(p_t)}
    =
    \exp\left(H_2(p_t)-H_1(p_t)\right)
    \in (0,1].
\end{equation}
Here $p_t$ denotes the next-token distribution of either a base or instruction-tuned model, and the range follows from the inequality $H_2(p_t)\leq H_1(p_t)$. The inequality becomes an equality if and only if the normalized distribution is uniform over its nonzero support. If this support contains $m$ tokens, then
\begin{equation}
H_1(p_t)=H_2(p_t)=\ln m,
\qquad
N_1(p_t)=N_2(p_t)=m,
\qquad
R(p_t)=1.
\end{equation}
Thus, $(N_1,R)$ jointly characterizes a plateau: $N_1$ identifies its order, while $R$ measures its proximity to uniformity. In particular, $(N_1,R)\approx(m,1)$ indicates an entropy-compressed state with roughly $m$ plausible tokens.

To examine whether the effective support ratio $R$ consistently reveals the multimodal structure of next-token distributions, we compute the joint distribution of $R$ and $N_1$ over assistant-token positions for both base and instruction-tuned models on UltraFeedback~\citep{cui2023ultrafeedback}, Magicoder-OSS-Instruct-75K~\citep{wei2023magicoder}, and NuminaMath-CoT~\citep{li2024numinamath}, which represent general-domain instruction following, code generation, and mathematical reasoning, respectively. We evaluate three model variants, Qwen3-4B~\citep{yang2025qwen3}, Llama-3.1-8B~\citep{grattafiori2024llama}, and Qwen3-30B-A3B~\citep{yang2025qwen3}, to assess whether the observed structure generalizes across model families, architectures, and scales. We compute $N_1$ and $N_2$ over the normalized top-$K$ distribution with $K=30$, from which $R$ is derived.

As shown in Figure~\ref{fig:plateau}, the joint distribution of $N_1$ and $R$ exhibits clear vertical ridges near integer values of $N_1 = 1, 2, 3, 4, 5, 10, \ldots$, especially in high-$R$ regions. These ridges indicate that the corresponding tokens lie in near-uniform entropy-compressed states with $H_1 \approx \ln m$, where $m$ denotes the effective number of plausible next-token alternatives. Importantly, these ridges persist beyond the previously emphasized low-order regime, suggesting that higher-order entropy-compressed states are widespread across architectures and datasets. The curved lower envelope corresponds to distributions dominated by a single token, with the remaining mass spread over a low-probability tail under finite top-$K$ normalization; Appendix~\ref{app:finite_topk_envelope} derives this boundary.

\subsection{Plateau-\texorpdfstring{$k$}{k} States: From Low-Order Branching to High-Order Ambiguity}
\label{sec:interpreting_plateau_k}

We define entropy-compressed states of order $k$ as plateau-$k$ states, captured by
\begin{equation}
    N_1 \in [k-0.3, k+0.3] \quad \text{and} \quad R>0.95.
\end{equation}
Figure~\ref{fig:plateau_k} compares the proportion of plateau-$k$ states between base and instruction-tuned models. Instruction tuning generally increases low-order plateaus ($k=1$--$5$), showing that post-training concentrates probability not only on a single token ($k=1$) but also onto small sets of plausible alternatives. In contrast, this trend weakens or reverses at higher orders ($k>5$), where a visible concentration appears around $k=10$. We use $k=5$ as a descriptive cutoff rather than a strict boundary, since semantic and structural ambiguity can occur at both low and high orders; Appendix~\ref{app:plateau_examples} provides qualitative examples.

\looseness=-1
\paragraph{Low-order plateau-$k$ states: logical branching.}
Low-order plateau states typically correspond to semantic or logical branching. In these cases, the model assigns comparable probabilities to a small set of locally plausible tokens, which may include the observed target token and other valid alternatives. Such states indicate that the next-token distribution preserves multiple plausible continuation paths rather than collapsing to one token.

\paragraph{High-order plateau-$k$ states: structural ambiguity.}
High-order plateau states exhibit qualitatively different behavior. They often arise from structural ambiguity, where the local context does not contain enough information to determine a unique next token. Typical cases include random IDs, dates, timestamps, and URLs. In such positions, the model distributes probability across alternative digits or characters; in particular, the concentration around $k=10$ largely comes from nearly uniform alternatives among the digits 0--9. This suggests that higher-order plateaus capture diffuse structural uncertainty, rather than the semantic branching observed in the low-order regime.

This observation highlights a key limitation of cross-entropy: it supervises only the target token, leaving preferences among plausible alternatives unconstrained and distorting uncertainty in both regimes. Plateau states clarify what is at stake: their non-label candidates are not tail noise, but continuations the pretrained model judges roughly as plausible as the target. Preserving them prevents fine-tuning from discarding endorsed alternatives, while still allowing the target to absorb probability mass. What deserves protection is therefore this organization rather than individual probabilities: the integer ridges in Figure~\ref{fig:plateau} show that the base model allocates uncertainty into discrete, near-uniform regimes matched to the number of admissible continuations. Preserving it does not mean freezing the entropy profile. Figure~\ref{fig:plateau_k} shows that instruction tuning legitimately sharpens low-order plateaus; the target may absorb arbitrary mass provided the relative arrangement of the remaining candidates survives. Only their indiscriminate destruction must be ruled out. This suggests a direct route to forgetting: by driving mass onto $y_t$, cross-entropy suppresses endorsed alternatives, making previously held continuations unreachable even as single-sample accuracy improves.\looseness=-1

\section{Method: Preserving Non-Label Relative Structure}
\label{sec:method}

Motivated by the observation that pretrained models encode structured relative preferences across both branching and diffuse uncertainty regimes, we propose \textbf{LP-SFT} (\textbf{L}ocal-\textbf{P}reserving \textbf{S}upervised \textbf{F}ine-\textbf{T}uning) with two design principles: \textbf{target-token removal} and \textbf{local normalization} of relative preferences (Figure~\ref{fig:lp_sft_overview}).

\subsection{A General Non-Label Preservation Objective}
\label{sec:general_objective}

Given an instruction-response pair with prompt $\mathbf{x}$ and target sequence $\mathbf{y}=(y_1, \dots, y_T)$, we denote the frozen base model distribution and the trainable model distribution at position $t$ by
$q_t(\cdot)=p_{\theta_0}(\cdot \mid \mathbf{x}, \mathbf{y}_{<t})$
and
$p_t(\cdot)=p_{\theta}(\cdot \mid \mathbf{x}, \mathbf{y}_{<t})$,
respectively.

We first select a candidate set $S_t$ from the frozen base-model distribution and remove the target token $y_t$ to form the preservation set $\mathcal{A}_t$:
\begin{equation}
    \label{eq:action_set}
    S_t = C_{K_{t}}(q_t),
\qquad \mathcal{A}_t = S_t \setminus \{y_t\}.
\end{equation}
Section~\ref{sec:interpreting_plateau_k} only imposes a lower bound on $K_t$: because preservation is probability-weighted, adding tokens beyond the plateau carries negligible weight, while setting $K_t$ too small truncates endorsed alternatives. We therefore use one generous budget, $K_t=K_{\max}=10$, covering the low-order plateaus and the regime around $k=10$, with no per-token schedule. Appendix~\ref{app:preservation_Kt} confirms both directions.

We then apply \textbf{local normalization}, restricting the base and trainable distributions to $\mathcal{A}_t$ and renormalizing within this non-label subset:
\begin{equation}
\label{eq:local_normalization}
\begin{aligned}
\hat{q}_t(v) &= \frac{q_t(v)}{\sum_{u\in \mathcal{A}_t}q_t(u)} = \frac{q_t(v)}{Q_t},
\qquad&
\hat{p}_t(v) &= \frac{p_t(v)}{\sum_{u\in \mathcal{A}_t}p_t(u)} = \frac{p_t(v)}{P_t},
\qquad \forall v \in \mathcal{A}_t ,
\end{aligned}
\end{equation}
where $Q_t = q_t(\mathcal{A}_t)$ and $P_t = p_t(\mathcal{A}_t)$ denote the total probability masses that the two models assign to the preservation set.
Here, $\hat{q}_t$ and $\hat{p}_t$ capture relative preferences over the non-target alternatives. Local normalization aligns naturally with this plateau structure: it preserves the relative probability profile among candidates where multimodal knowledge resides, while leaving their total mass $P_t$ unconstrained so that the supervised target $y_t$ can absorb probability freely. In contrast, matching unnormalized probabilities directly constrains total non-target mass and opposes cross-entropy, as Section~\ref{sec:gradient_view} demonstrates. The preservation loss is defined as
\begin{equation}
\label{eq:lp_loss}
\mathcal{L}_{\mathrm{LP}}^t
=
D_{\mathrm{KL}}\!\left(\hat{q}_t \,\|\, \hat{p}_t\right)
=
\sum_{v\in\mathcal{A}_t}
\hat{q}_t(v)
\ln
\frac{\hat{q}_t(v)}
{\hat{p}_t(v)},
\end{equation}
and the full training objective is
\begin{equation}
\label{eq:full_objective}
\mathcal{L}(\theta)
=
\sum_{t=1}^{T}
\left[
-\ln p_t(y_t)
+
\mu \, \mathcal{L}_{\mathrm{LP}}^t
\right],
\end{equation}
where $\mu$ controls the preservation strength. Since $\hat{q}_t$ is derived from the frozen base model, it is a fixed target, so minimizing Eq.~\eqref{eq:lp_loss} is equivalent, up to a constant independent of $\theta$, to minimizing the local cross-entropy $-\sum_{v \in \mathcal{A}_t} \hat{q}_t(v) \ln \hat{p}_t(v)$.

Unlike standard distillation~\citep{hinton2015distilling,gu2024minillm,agarwal2024policy}, which transfers knowledge by matching a teacher distribution, LP-SFT uses the frozen base model only to preserve the local structure of non-label alternatives.

\subsection{A Gradient View: Why Both Principles Matter}
\label{sec:gradient_view}

\subsubsection{Local Normalization}

The locally normalized loss $\mathcal{L}_{\mathrm{LP}}^t$ has the gradient (Appendix~\ref{app:grad_normalization})
\begin{equation}
\label{eq:normalized_gradient_cases}
\frac{\partial\mathcal{L}_{\mathrm{LP}}^t}{\partial z_t(w)}
=
\begin{cases}
\dfrac{p_t(w)}{P_t}-\dfrac{q_t(w)}{Q_t}, & w\in\mathcal{A}_t,\\
0, & w\notin\mathcal{A}_t.
\end{cases}
\end{equation}
Thus, it only adjusts the relative logits within $\mathcal{A}_t$, leaving those outside unaffected. In contrast, the unnormalized truncation acts fundamentally differently, as seen from its exact decomposition:
\begin{equation}
\label{eq:unnorm_decomp}
\widetilde{\mathcal{L}}^t = \sum_{v\in\mathcal{A}_t} q_t(v)\ln\frac{q_t(v)}{p_t(v)} = Q_t\mathcal{L}_{\mathrm{LP}}^t + Q_t\ln\frac{Q_t}{P_t}.
\end{equation}

The principal difference is the mass term $Q_t\ln(Q_t/P_t)$, which is strictly decreasing in $P_t$ and therefore exerts a one-sided pressure toward $P_t\to1$, rather than matching the preservation-set mass to its base-model value. This pressure is also visible from the full gradient:
\begin{equation}
\label{eq:unnorm_gradient_cases}
\frac{\partial\widetilde{\mathcal{L}}^t}{\partial z_t(w)}
=
\begin{cases}
Q_t p_t(w)-q_t(w), & w\in\mathcal{A}_t,\\
Q_t p_t(w), & w\notin\mathcal{A}_t.
\end{cases}
\end{equation}
Every logit outside $\mathcal{A}_t$ therefore has a positive gradient, while the gradients inside satisfy $\sum_{w\in\mathcal{A}_t}\partial\widetilde{\mathcal{L}}^t/\partial z_t(w)=Q_t(P_t-1)<0$. Gradient descent thus suppresses logits outside the preservation set and raises those inside as a group, shifting probability mass into $\mathcal{A}_t$. In addition, the factor $Q_t$ scales the relative-preservation term in Eq.~\eqref{eq:unnorm_decomp}, making the effective preservation weight $\mu Q_t$ vary across token positions. Local normalization removes both this mass-shifting pressure and the position-dependent reweighting, leaving only the KL constraint on the relative distributions within $\mathcal{A}_t$.\looseness=-1

\subsubsection{Target-Token Removal}
Target-token removal eliminates a direct conflict at the supervised logit. If $y_t$ is retained in a locally normalized preservation set, let $\mathcal{L}_{\mathrm{pres}}^t$ denote the resulting preservation term. The target-logit gradients of cross-entropy and the preservation term are
\[
\frac{\partial \mathcal{L}_{\mathrm{CE}}^t}{\partial z_t(y_t)}
=p_t(y_t)-1,
\qquad
\frac{\partial \mathcal{L}_{\mathrm{pres}}^t}{\partial z_t(y_t)}
=\hat{p}_t(y_t)-\hat{q}_t(y_t).
\]
Cross-entropy increases the target probability, whereas the preservation term pulls the target logit back once $\hat{p}_t(y_t)$ exceeds its base-model value $\hat{q}_t(y_t)$. Denoting the stationary target probability under cross-entropy plus preservation by $p_{t,\mathrm{pres}}^{\star}(y_t)$, we obtain, for any non-degenerate preservation set,
\begin{equation}
\label{eq:target_cap}
p_{t,\mathrm{pres}}^{\star}(y_t)
\le
\frac{1+\mu\hat{q}_t(y_t)}
{1+\mu}
<1 .
\end{equation}
By combining local normalization with target-token removal, LP-SFT guarantees that the target-logit fixed point remains at the cross-entropy optimum, $p_t^{\star}(y_t)=1$. This does not imply that the trained model assigns probability one to every supervised token; rather, it means that LP-SFT preserves non-label structure without attenuating SFT's direct supervision on $y_t$. Appendix~\ref{app:grad_target} derives the exact stationary points for all four combinations, and Appendix~\ref{app:preservation_set_ablation} verifies their empirical ordering.

\begin{table*}[t]
\centering
\scriptsize
\setlength{\tabcolsep}{5pt}
\renewcommand{\arraystretch}{1.0}
\caption{
\textbf{Fine-tuning results on the mixed-domain dataset.}
LP-SFT achieves the strongest result among fine-tuned methods on every reported metric.
}
\label{tab:allmix_results}

\begin{tabular}{@{}ll cccc cccc c c@{}}
\toprule
\multirow{3}{*}{\textbf{Model}}
& \multirow{3}{*}{\textbf{Method}}
& \multicolumn{4}{c}{\textbf{Math}}
& \multicolumn{4}{c}{\textbf{Code}}
& \multicolumn{1}{c}{\textbf{GEN.}}
& \multirow{2}{*}{\textbf{Avg.}} \\
\cmidrule(lr){3-6}
\cmidrule(lr){7-10}
\cmidrule(lr){11-11}
&
& \multicolumn{2}{c}{\makecell{\textbf{MATH-500}\\pass@1 / pass@16}}
& \multicolumn{2}{c}{\makecell{\textbf{AIME 21--26}\\pass@1 / pass@32}}
& \multicolumn{2}{c}{\makecell{\textbf{MBPP+}\\pass@1 / pass@10}}
& \multicolumn{2}{c}{\makecell{\textbf{HE+}\\pass@1 / pass@10}}
& \makecell{\textbf{MMLU}\\5-shot Acc.}
& \makecell{ \\ All} \\
\midrule

\multirow{7}{*}{Qwen3-14B}
& Base  & 69.50 & 89.40 & 7.00 & 26.11 & 69.62 & 87.88 & 77.74 & 93.32 & 80.51 & 66.79 \\
\cmidrule(lr){2-12}
& CE    & 51.60 & 85.80 & 4.62 & 21.67 & 57.12 & 81.49 & 58.26 & 88.49 & 77.92 & 58.55 \\
& DFT   & 40.12 & 62.20 & 3.25 & 10.56 & 68.16 & 70.84 & 65.95 & 71.48 & 77.90 & 52.27 \\
& EAFT  & 51.00 & 85.60 & 4.77 & 26.11 & 61.15 & 82.92 & 64.88 & 89.24 & 78.44 & 60.46 \\
& GEM   & 47.89 & 85.80 & 3.99 & 22.22 & 54.29 & 82.63 & 55.79 & 86.26 & 77.53 & 57.38 \\
& ASFT  & 54.57 & 82.80 & 4.10 & 21.11 & 68.07 & 78.18 & 72.10 & 83.20 & 78.61 & 60.30 \\
\rowcolor{BAShade} \cellcolor{white}
& LP-SFT & \textbf{64.20} & \textbf{89.20} & \textbf{6.79} & \textbf{27.78} & \textbf{69.17} & \textbf{85.74} & \textbf{74.66} & \textbf{90.25} & \textbf{80.05} & \textbf{65.32} \\
\bottomrule
\end{tabular}
\end{table*}

\begin{figure*}[t]
    \centering
    \includegraphics[width=\textwidth]{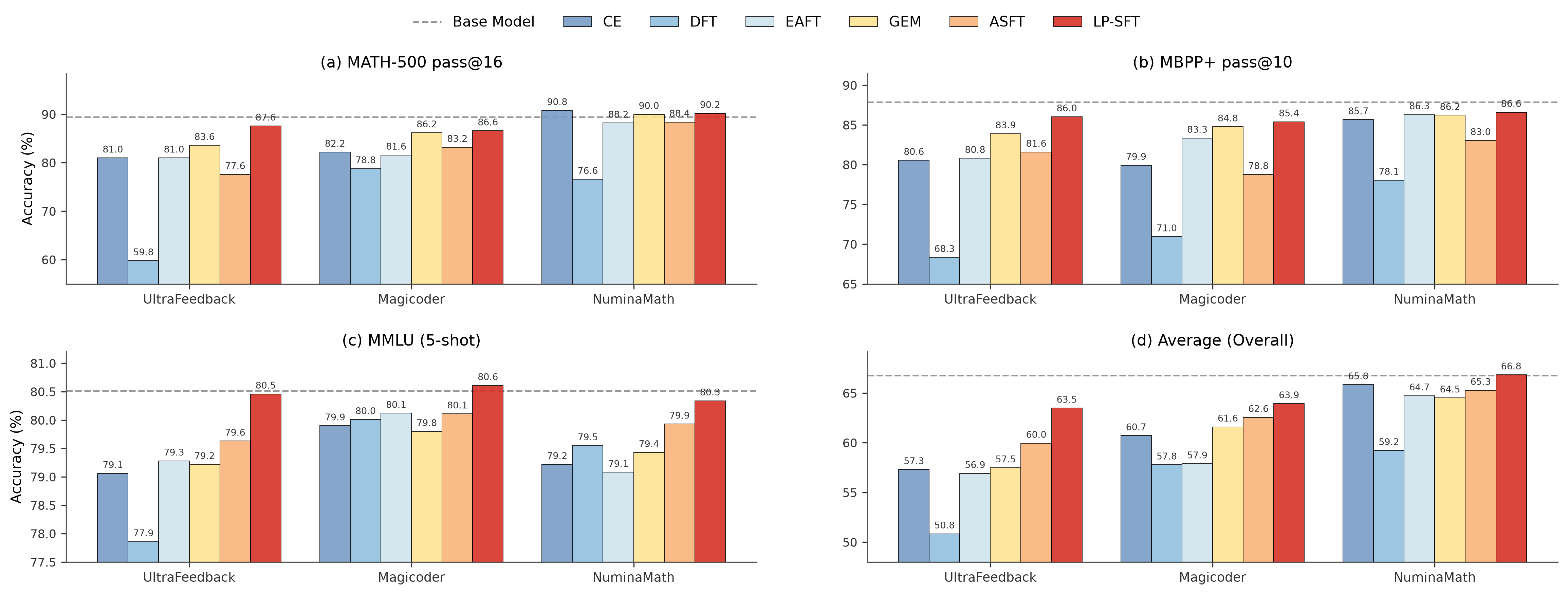}
    \caption{
    \textbf{Qwen3-14B fine-tuning results across three training datasets.}
    LP-SFT achieves the highest overall average on UltraFeedback, Magicoder, and NuminaMath, while consistently preserving strong pass@$k$ performance and general knowledge.
    }
    \label{fig:qwen14b_results}
\end{figure*}

\subsection{Efficient Implementation}
\label{sec:efficient_impl}

A naive implementation of LP-SFT would require keeping both the frozen base model and the trainable model in memory and running an additional base-model forward pass at each training step. Since the base model $\theta_0$ is frozen and the training data is fixed, we instead precompute the required base-distribution information offline.

As shown in Algorithm~\ref{alg:lpsft} in Appendix~\ref{app:algorithm}, LP-SFT is implemented in two stages. In Stage 1, we perform a single offline pass over the training dataset with the base model and cache the top-$K_{\max}$ candidate logits required to construct the local preservation set. In Stage 2, fine-tuning proceeds with only the standard forward pass of the trainable model. The local preservation loss $\mathcal{L}_{\mathrm{LP}}^t$ is computed at each training step using the cached base logits and the current model logits restricted to $\mathcal{A}_t$.

\section{Experiments} 
\label{sec:experiments} 
 
We evaluate \textbf{LP-SFT} across multiple model scales and domains, focusing on two questions: (1) how it balances single-sample reliability, measured by pass@1, with finite-budget solution accessibility, measured by pass@$k$, and (2) whether it mitigates catastrophic forgetting while improving overall performance over standard SFT and recent alternatives.

\subsection{Setup} 
 
\paragraph{Models and datasets.} 
Our main experiments evaluate three base models spanning different scales: \textbf{Qwen3-4B}, \textbf{Llama-3.1-8B}, and \textbf{Qwen3-14B}. To test domain learning and capability preservation, we use \textbf{Magicoder-OSS-Instruct-75K} for code, a 100K subset of \textbf{NuminaMath-CoT} for mathematics, and \textbf{UltraFeedback} (61K) for general alignment. We additionally conduct a rank-64 LoRA scaling study on \textbf{Qwen2.5-72B} with UltraFeedback-30K in Appendix~\ref{app:qwen25_72b_results}.
 
\paragraph{Baselines.}
We compare LP-SFT with standard supervised fine-tuning using cross-entropy loss (\textbf{CE}), Dynamic Fine-Tuning (\textbf{DFT})~\citep{wu2025generalization}, Entropy-Adaptive Fine-Tuning (\textbf{EAFT})~\citep{diao2026entropy}, \textbf{GEM}~\citep{li2025preserving}, and Anchored Supervised Fine-Tuning (\textbf{ASFT})~\citep{zhu2025anchored}. For LP-SFT, we set the preservation weight to $\mu=1$ and use a fixed top-$K_{\max}$ preservation set with $K_{\max}=10$ by default.

\paragraph{Evaluation benchmarks and metrics.}
We evaluate fine-tuned models across mathematics (\textbf{MATH-500}, \textbf{AIME 2021--2026}), coding (\textbf{MBPP+}, \textbf{HumanEval+}), and general knowledge (\textbf{MMLU}, 5-shot)~\citep{hendrycks2020measuring,hendrycks2021measuring,liu2023your}; detailed task setups are in Appendix~\ref{app:training_eval_details}. For generation tasks, we report both \textbf{pass@1} (single-sample reliability) and \textbf{pass@$k$} (finite-budget solution accessibility)~\citep{chen2021evaluating}. This pairing mirrors the mechanism in Section~\ref{sec:interpreting_plateau_k}: cross-entropy can maintain pass@1 on the supervised target, but by suppressing endorsed alternatives it makes other reasoning paths harder to reach, which surfaces as a decline in pass@$k$. Avg.\ denotes the mean of all displayed metrics.

\begin{figure}[t]
    \centering
    \vspace{-2mm}
    \includegraphics[width=\linewidth]{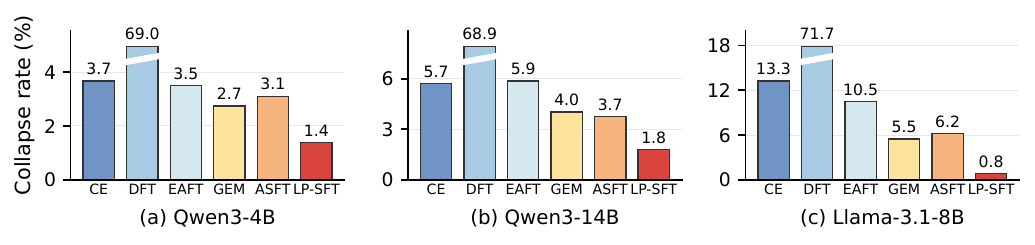}
    \vspace{-5mm}
    \caption{
    \textbf{Collapse of base-endorsed alternatives after fine-tuning on UltraFeedback.}
    Collapse rate (Eq.~\ref{eq:collapse_rate}; lower is better) on base-generated MATH-500 solutions (App.~\ref{app:collapse_diag}); DFT clipped.
    }
    \label{fig:alt_collapse}
    \vspace{-5mm}
\end{figure}

\subsection{Mixed-Domain Fine-Tuning}

We first evaluate LP-SFT in a mixed-domain setting, in which Qwen3-14B is fine-tuned on a mixture of three datasets covering mathematical reasoning, code generation, and general instruction following, so that competing optimization signals can exacerbate forgetting. As shown in Table~\ref{tab:allmix_results}, LP-SFT achieves the best results across all displayed metrics among the fine-tuned models, including both pass@1 and pass@$k$.

The distinction is especially clear in general-knowledge retention. Every competing fine-tuning objective substantially reduces MMLU accuracy from the base model's 80.51 to around 78, whereas LP-SFT retains 80.05, a drop of only 0.46 points. The same pattern holds in aggregate performance: competing methods fall 6.33--14.52 average points below the base model, whereas LP-SFT reduces this gap to only 1.47 points.\looseness=-1

\subsection{Domain-Specific Fine-Tuning}

We further evaluate LP-SFT under domain-specific fine-tuning on three datasets: UltraFeedback, Magicoder-OSS-Instruct-75K, and NuminaMath-CoT. Figure~\ref{fig:qwen14b_results} summarizes the Qwen3-14B results across these datasets, while Table~\ref{tab:ultrafeedback_results} reports detailed UltraFeedback results for Qwen3-4B, Qwen3-14B, and Llama-3.1-8B. Additional results on Magicoder and NuminaMath are provided in Appendix~\ref{app:additional_results}.

Across all combinations of the three base models and training datasets, LP-SFT achieves the highest overall average. The gains are most pronounced on UltraFeedback, where general-domain fine-tuning can substantially degrade mathematical reasoning and code generation. On Magicoder and NuminaMath, LP-SFT also improves the overall average and MMLU accuracy while maintaining competitive task-specific pass@$k$.

\begin{table*}[t]
\centering
\scriptsize
\setlength{\tabcolsep}{5pt}
\renewcommand{\arraystretch}{1.0}
\caption{
\textbf{Fine-tuning results on UltraFeedback.}
LP-SFT obtains the highest overall average for all three model families, with consistent gains in MBPP+ pass@10 and MMLU across all models, alongside substantial mathematical improvements on Qwen3.
}
\label{tab:ultrafeedback_results}

\begin{tabular}{@{}ll cccc cccc c c@{}}
\toprule
\multirow{3}{*}{\textbf{Model}}
& \multirow{3}{*}{\textbf{Method}}
& \multicolumn{4}{c}{\textbf{Math}}
& \multicolumn{4}{c}{\textbf{Code}}
& \multicolumn{1}{c}{\textbf{GEN.}}
& \multirow{2}{*}{\textbf{Avg.}} \\
\cmidrule(lr){3-6}
\cmidrule(lr){7-10}
\cmidrule(lr){11-11}
&
& \multicolumn{2}{c}{\makecell{\textbf{MATH-500}\\pass@1 / pass@16}}
& \multicolumn{2}{c}{\makecell{\textbf{AIME 21--26}\\pass@1 / pass@32}}
& \multicolumn{2}{c}{\makecell{\textbf{MBPP+}\\pass@1 / pass@10}}
& \multicolumn{2}{c}{\makecell{\textbf{HE+}\\pass@1 / pass@10}}
& \makecell{\textbf{MMLU}\\5-shot Acc.}
& \makecell{ \\ All} \\
\midrule

\multirow{6}{*}{Qwen3-4B}
& CE    & 38.66 & 79.20 & 1.27 & 16.67 & 51.28 & 72.57 & 62.26 & 86.33 & 71.56 & 53.31 \\
& DFT   & 37.75 & 59.60 & 0.75 & 5.00 & 54.62 & 61.67 & 64.42 & 68.62 & 70.64 & 47.01 \\
& EAFT  & 26.90 & 75.60 & 0.90 & 13.33 & 52.18 & 76.85 & 62.29 & 86.54 & 72.23 & 51.87 \\
& GEM   & 36.39 & 79.00 & 1.09 & 15.56 & 46.97 & 75.85 & 56.07 & 87.35 & 71.38 & 52.18 \\
& ASFT  & 40.86 & 74.00 & 1.74 & 12.78 & 57.46 & 71.37 & 68.08 & 83.72 & 71.92 & 53.55 \\
\rowcolor{BAShade} \multirow{-6}{*}{\cellcolor{white}Qwen3-4B}
& LP-SFT & \textbf{50.15} & \textbf{83.40} & \textbf{3.39} & \textbf{19.44} & \textbf{57.79} & \textbf{80.69} & \textbf{69.30} & \textbf{92.29} & \textbf{72.62} & \textbf{58.79} \\
\midrule

\multirow{6}{*}{Qwen3-14B}
& CE    & 45.16 & 81.00 & 1.79 & 16.11 & 60.32 & 80.56 & 64.94 & 86.84 & 79.06 & 57.31 \\
& DFT   & 40.27 & 59.80 & 0.75 & 4.44 & 63.81 & 68.35 & 68.08 & 74.17 & 77.86 & 50.84 \\
& EAFT  & 43.00 & 81.00 & 1.96 & 15.56 & 59.38 & 80.81 & 63.87 & 87.39 & 79.28 & 56.92 \\
& GEM   & 43.00 & 83.60 & 2.14 & 17.78 & 56.75 & 83.90 & 61.37 & 89.62 & 79.22 & 57.49 \\
& ASFT  & 46.06 & 77.60 & 2.14 & 16.67 & \textbf{71.01} & 81.59 & \textbf{77.74} & 87.23 & 79.63 & 59.96 \\
\rowcolor{BAShade} \multirow{-6}{*}{\cellcolor{white}Qwen3-14B}
& LP-SFT & \textbf{54.71} & \textbf{87.60} & \textbf{4.70} & \textbf{23.33} & 68.00 & \textbf{86.03} & 74.30 & \textbf{92.31} & \textbf{80.46} & \textbf{63.49} \\
\midrule

\multirow{6}{*}{Llama-3.1-8B}
& CE    & \textbf{10.76} & \textbf{45.80} & \textbf{0.23} & \textbf{6.11} & 31.61 & 51.04 & 33.05 & 59.12 & 61.11 & 33.20 \\
& DFT   & 5.16 & 16.60 & 0.21 & 5.00 & 26.59 & 32.55 & 25.27 & 32.18 & 55.11 & 22.07 \\
& EAFT  & 9.31 & 43.40 & 0.19 & 5.56 & 27.08 & 49.91 & \textbf{34.15} & 59.50 & 61.14 & 32.25 \\
& GEM   & 8.82 & 44.00 & 0.21 & \textbf{6.11} & 30.34 & 54.33 & 32.44 & \textbf{62.84} & 61.00 & 33.34 \\
& ASFT  & 8.45 & 35.60 & 0.05 & 1.67 & 36.59 & 55.49 & 31.80 & 51.07 & 60.58 & 31.26 \\
\rowcolor{BAShade} \multirow{-6}{*}{\cellcolor{white}Llama-3.1-8B}
& LP-SFT & 9.41 & 42.00 & 0.12 & 3.33 & \textbf{37.00} & \textbf{63.04} & 29.79 & 57.75 & \textbf{62.78} & \textbf{33.91} \\
\bottomrule
\end{tabular}

\end{table*}

\paragraph{Comparison with baselines.}
Figure~\ref{fig:qwen14b_results} and Table~\ref{tab:ultrafeedback_results} show that DFT, ASFT, and GEM each sacrifice pass@1 or pass@$k$, whereas LP-SFT balances both and preserves general knowledge. Figure~\ref{fig:alt_collapse} traces these trade-offs to base-endorsed alternatives $\mathcal{E}_t$ via the collapse rate
\begin{equation}
\label{eq:collapse_rate}
\mathrm{CR}=\frac{\sum_t\sum_{v\in\mathcal{E}_t}\mathbf{1}[p_t(v)<10^{-3}]}{\sum_t|\mathcal{E}_t|},\qquad
\mathcal{E}_t=\{v\in\mathcal{A}_t:q_t(v)\ge0.05\}.
\end{equation}
CE collapses 3.7--13.3\% of them and DFT about 70\%, whereas LP-SFT reduces CR to 0.8--1.8\%. Since LP-SFT constrains neither $y_t$ nor the alternative mass $P_t$, this suggests that keeping endorsed alternatives from degenerating is a key source of its gain. Gains are largest on Qwen3, whereas CE remains stronger on Llama-3.1-8B math, suggesting that stronger bases offer more reliable structure.

\paragraph{Scaling and ablations.}
On Qwen2.5-72B, LP-SFT achieves the best aggregate score among all fine-tuned methods and slightly exceeds the base model, at CE-level fine-tuning cost plus a one-time cache (Tables~\ref{tab:app_qwen25_72b_results} and~\ref{tab:app_qwen25_72b_training_time}). Dropping either target-token removal or local normalization substantially weakens the gains (Appendix~\ref{app:preservation_set_ablation}); local top-$K$ preservation matches the full-vocabulary variant without a frozen reference model (Appendix~\ref{app:preservation_Kt}); and LP-SFT is robust to $\mu$ (Figure~\ref{fig:mu_ablation}).

\section{Conclusion}

Our entropy analysis shows that base-model predictions exhibit structured plateau states, indicating that useful knowledge resides in relative preferences among non-label alternatives rather than their absolute probability mass. This motivates two core design principles: target-token removal and local normalization, which jointly safeguard non-label structure without restricting the supervised target. Crucially, local top-$K$ preservation matches full-vocabulary retention without reference-model overhead. Across diverse domains and model scales up to Qwen2.5-72B, LP-SFT consistently achieves the strongest balance between single-sample reliability and finite-budget solution accessibility.\looseness=-1

\subsection*{AI use statement}

Generative AI tools were used throughout this research. Tasks for which
disclosure is required include proposing and refining hypotheses, developing
the conceptual framework, designing the methodology and experiments,
formulating mathematical claims and assisting in writing the proofs in the
appendices, implementing the training, evaluation, and entropy-analysis code,
preparing and subsampling the datasets, analyzing the qualitative token-level
examples, interpreting experimental results, and translating author-written
text into English. We did not use generative AI tools to generate synthetic
datasets: all training and evaluation data come from existing public datasets
and benchmarks. Transcribing recordings of research material and formulating
survey or interview questions are not applicable to this work.

We additionally used generative AI tools for tasks in the recommended
disclosure category, including drafting and editing the manuscript, creating
figures and \LaTeX{} tables, brainstorming candidate titles, and searching for
related literature.

We have reviewed all AI-assisted work. The authors take responsibility for the
final content of this work, including all text, claims, and artifacts produced
with the aid of generative AI.

\subsection*{Reproducibility statement}
The method and its optimization objective are specified in
Section~\ref{sec:method}, with implementation pseudocode provided in
Appendix~\ref{app:algorithm}. Appendix~\ref{app:training_eval_details}
documents the training and evaluation protocols, and
Appendix~\ref{app:ablation_efficiency} reports hyperparameter sensitivity,
component ablations, and computational costs. Complete derivations supporting
the theoretical analysis are also included in the appendix.

\bibliographystyle{plain}
\bibliography{references}

\appendix

\section{Derivation of the Finite-top-\texorpdfstring{$K$}{K} Lower Envelope}
\label{app:finite_topk_envelope}

We derive the finite-top-$K$ lower-envelope branch of $R$ for a given Shannon effective support size $N_1$. Let $\hat{p}=(\hat{p}_1,\dots,\hat{p}_K)$ denote the normalized top-$K$ next-token distribution. For a fixed $N_1$, minimizing $R$ is equivalent to minimizing $N_2$, or equivalently maximizing the collision probability $\sum_{i=1}^{K}\hat{p}_i^2$, under a fixed Shannon entropy constraint:
\begin{equation}
\label{eq:lower_envelope_optimization}
\begin{aligned}
    \max_{\hat{p}} \quad
    & \sum_{i=1}^{K}\hat{p}_i^2 \\
    \mathrm{s.t.} \quad
    &
    \left\{
    \begin{aligned}
        &\sum_{i=1}^{K}\hat{p}_i=1, \\
        &-\sum_{i=1}^{K}\hat{p}_i\ln \hat{p}_i=\ln N_1, \\
        &\hat{p}_i\ge 0,\qquad i=1,\dots,K.
    \end{aligned}
    \right.
\end{aligned}
\end{equation}

The Lagrangian of this constrained optimization problem is
\begin{equation}
    \mathcal{L}
    =
    \sum_{i=1}^{K}\hat{p}_i^2
    -
    \lambda\left(\sum_{i=1}^{K}\hat{p}_i-1\right)
    -
    \mu\left(
    -\sum_{i=1}^{K}\hat{p}_i\ln \hat{p}_i
    -
    \ln N_1
    \right).
\end{equation}
For any positive component $\hat{p}_i>0$, the stationarity condition gives
\begin{equation}
    \frac{\partial \mathcal{L}}{\partial \hat{p}_i}
    =
    2\hat{p}_i-\lambda+\mu(\ln \hat{p}_i+1) = 0,
\end{equation}
or equivalently
\begin{equation}
\label{eq:kkt_two_values}
    2\hat{p}_i+\mu\ln \hat{p}_i=C,
\end{equation}
where $C=\lambda-\mu$ is a constant independent of $i$. Since the scalar equation $2x+\mu\ln x=C$ has at most two positive solutions, any interior stationary extremum has at most two distinct positive probability values. Boundary extrema are covered by allowing some probabilities to be zero.

Therefore, the general two-level extremal family can be written as
\begin{equation}
\label{eq:two_level_general}
    \hat{p}^{\dagger}(j,m,a,b)
    =
    \Big(
        \underbrace{a,\ldots,a}_{j\text{ entries}},\;
        \underbrace{b,\ldots,b}_{m-j\text{ entries}},\;
        0,\ldots,0
    \Big),
\end{equation}
where
\begin{equation}
    a>b\ge 0,
    \qquad
    1\le j\le m-1,
    \qquad
    2\le m\le K.
\end{equation}
The values $(a,b)$ are determined by the normalization and entropy constraints
\begin{equation}
    ja+(m-j)b=1,
    \qquad
    -ja\ln a-(m-j)b\ln b=\ln N_1.
\end{equation}

Among these two-level candidates, the lower-envelope branch observed in the $(N_1,R)$ plane is attained by the most asymmetric case, where the larger probability value appears once. Intuitively, concentrating the high-probability mass on a single coordinate increases the collision probability, while the remaining mass forms a uniform residual tail. Enumeration over the finite range used in our analysis, with $m\in\{2,\dots,30\}$ and $N_1\in(1,m]$, confirms that the lower branch is attained by $j=1$.

Thus, the relevant extremal family reduces to the one-dominant / uniform-tail form
\begin{equation}
\label{eq:extremal}
    \hat{p}^{\star}(q, m)
    =
    \Big(
        q,\;
        \underbrace{\tfrac{1-q}{m-1}, \ldots, \tfrac{1-q}{m-1}}_{m-1 \text{ entries}},\;
        0,\ldots,0
    \Big),
    \qquad
    q\in[1/m,1],
    \quad
    m\in\{2,\dots,K\}.
\end{equation}

For this family, the Shannon entropy is
\begin{equation}
\label{eq:H_q_m}
    H_1(q,m)
    =
    -q\ln q
    -
    (1-q)\ln\frac{1-q}{m-1},
\end{equation}
and the collision probability is
\begin{equation}
\label{eq:collision_q_m}
    \sum_{i=1}^{K}\hat{p}_i^2
    =
    q^2
    +
    \frac{(1-q)^2}{m-1}.
\end{equation}

The finite-top-$K$ lower-envelope branch can then be written as
\begin{equation}
    R_{\min}(N_1;K)
    =
    \min_{\substack{
        2\le m\le K,\; q\in[1/m,1]
    }}
    \frac{1}{
    N_1
    \left[
    q^2+
    \frac{(1-q)^2}{m-1}
    \right]
    }
    \quad
    \mathrm{s.t.}
    \quad
    H_1(q,m)=\ln N_1 .
\end{equation}
Infeasible values of $m$ are automatically excluded by the entropy constraint $H_1(q,m)=\ln N_1$. In our entropy analyses, $K=30$, since both $N_1$ and $N_2$ are computed from the normalized top-$K$ distribution with $K=30$.

\section{Representative Plateau Examples}
\label{app:plateau_examples}

We provide representative token-level examples for different plateau orders. Each example shows a truncated context window around the ground-truth target token, with the target highlighted, together with the corresponding prediction distribution. For $k\leq 10$, we show the top-10 candidates; for $k>10$, we show the top-20 candidates.

\captionsetup{hypcap=false}

\paragraph{Instance 1: $k=2$}

\noindent
\textbf{Target Token:} \texttt{of}
\hspace{2em}
\textbf{$N_1$:} $2.000$
\hspace{2em}
\textbf{$R$:} $1.0000$
\begin{tcolorbox}[
    colback=backcolour,
    colframe=framecolour,
    title=\textbf{Context Window (Truncated)},
    coltitle=black!80,
    fonttitle=\sffamily\small,
    sharp corners,
    boxrule=0.8pt,
    left=4pt, right=4pt, top=2pt, bottom=2pt
]
\begin{lstlisting}[basicstyle=\ttfamily\small]
To solve for \( x \), we first combine like terms:

\[ \frac{7}{2}x = 14 \]

Now, we multiply both sides @\target{of}@ the equation by \( \frac{2}{7} \) to isolate \( x \):

\[ x = 14 \times \frac{2}{7} \]
\end{lstlisting}
\end{tcolorbox}

\begin{center}
    \begin{minipage}{\linewidth}
        \centering
        \small
        \setlength{\tabcolsep}{14pt} 
        \captionof{table}{Top-10 Prediction Distribution for Instance 1 ($k=2$).} 
        \label{tab:instance_k2_prob}
        \vspace{-0.3em}
        
        \begin{tabular}{ccc@{\hskip 30pt}ccc} 
        \toprule
        \textbf{Rank} & \textbf{Token} & \textbf{Prob.} & 
        \textbf{Rank} & \textbf{Token} & \textbf{Prob.} \\
        \midrule
        1 & \texttt{of} & 0.500 & 6  & \texttt{by} & 0.000 \\
        2 & \texttt{by} & 0.500 & 7  & \texttt{By} & 0.000 \\
        3 & \texttt{with} & 0.000 & 8  & \texttt{of} & 0.000 \\
        4 & \texttt{cua} & 0.000 & 9  & \texttt{the} & 0.000 \\
        5 & \texttt{to} & 0.000 & 10 & \texttt{Thai-of} & 0.000 \\
        \bottomrule
        \end{tabular}
    \end{minipage}
\end{center}
\vspace{0.5em}

\paragraph{Instance 2: $k=3$}
\noindent
\textbf{Target Token:} \texttt{divide}
\hspace{2em}
\textbf{$N_1$:} $3.000$
\hspace{2em}
\textbf{$R$:} $0.9999$
\begin{tcolorbox}[
    colback=backcolour,
    colframe=framecolour,
    title=\textbf{Context Window (Truncated)},
    coltitle=black!80,
    fonttitle=\sffamily\small,
    sharp corners,
    boxrule=0.8pt,
    left=4pt, right=4pt, top=2pt, bottom=2pt
]
\begin{lstlisting}[basicstyle=\ttfamily\small]
7000(1 + r/1)^(1*2)
19828.80 = 17000(1 + r)^2

Now, @\target{divide}@ both sides by 17000:

19828.80 / 17000 = (1 + r)^2
\end{lstlisting}
\end{tcolorbox}

\begin{center}
    \begin{minipage}{\linewidth}
        \centering
        \small
        \setlength{\tabcolsep}{14pt} 
        \captionof{table}{Top-10 Prediction Distribution for Instance 2 ($k=3$).} 
        \label{tab:instance_k3_prob}
        \vspace{-0.3em}
        
        \begin{tabular}{ccc@{\hskip 30pt}ccc} 
        \toprule
        \textbf{Rank} & \textbf{Token} & \textbf{Prob.} & 
        \textbf{Rank} & \textbf{Token} & \textbf{Prob.} \\
        \midrule
        1 & \texttt{divide} & 0.333 & 6  & \texttt{dividing} & 0.000 \\
        2 & \texttt{let}    & 0.333 & 7  & \texttt{simplify} & 0.000 \\
        3 & \texttt{we}     & 0.333 & 8  & \texttt{to} & 0.000 \\
        4 & \texttt{solve}  & 0.000 & 9  & \texttt{divided} & 0.000 \\
        5 & \texttt{divide} & 0.000 & 10 & \texttt{square} & 0.000 \\
        \bottomrule
        \end{tabular}
    \end{minipage}
\end{center}
\vspace{0.5em}

\paragraph{Instance 3: $k=4$}

\noindent
\textbf{Target Token:} \texttt{+}
\hspace{2em}
\textbf{$N_1$:} $4.029$
\hspace{2em}
\textbf{$R$:} $0.9945$
\begin{tcolorbox}[
    colback=backcolour,
    colframe=framecolour,
    title=\textbf{Context Window (Truncated)},
    coltitle=black!80,
    fonttitle=\sffamily\small,
    sharp corners,
    boxrule=0.8pt,
    left=4pt, right=4pt, top=2pt, bottom=2pt
]
\begin{lstlisting}[basicstyle=\ttfamily\small]
+ 1 - \frac{3}{ax} \\
&= \frac{2a^2x^2 + ax - 3}{ax} \\
&= \frac{(2ax@\target{+}@3)(ax-1)}{ax}.
\end{aligned}
\]

For \(x=1\) to be a critical point, \(f'(1)=0\). Plugging in \(
\end{lstlisting}
\end{tcolorbox}

\begin{center}
    \begin{minipage}{\linewidth}
        \centering
        \small
        \setlength{\tabcolsep}{14pt} 
        \captionof{table}{Top-10 Prediction Distribution for Instance 3 ($k=4$).} 
        \label{tab:instance_k4_prob}
        \vspace{-0.3em}
        
        \begin{tabular}{ccc@{\hskip 30pt}ccc} 
        \toprule
        \textbf{Rank} & \textbf{Token} & \textbf{Prob.} & 
        \textbf{Rank} & \textbf{Token} & \textbf{Prob.} \\
        \midrule
        1 & \texttt{\textvisiblespace -} & 0.250 & 6  & \texttt{\textasciicircum} & 0.000 \\
        2 & \texttt{-}                   & 0.250 & 7  & \texttt{+\textbackslash}  & 0.000 \\
        3 & \texttt{+}                   & 0.250 & 8  & \texttt{-\textbackslash}  & 0.000 \\
        4 & \texttt{\textvisiblespace +} & 0.250 & 9  & \texttt{+a}               & 0.000 \\
        5 & \texttt{)\textasciicircum}   & 0.000 & 10 & \texttt{\_}               & 0.000 \\
        \bottomrule
        \end{tabular}
    \end{minipage}
\end{center}
\vspace{0.5em}

\paragraph{Instance 4: $k=5$}
\noindent
\textbf{Target Token:} \texttt{7}
\hspace{2em}
\textbf{$N_1$:} $5.277$
\hspace{2em}
\textbf{$R$:} $0.9647$
\begin{tcolorbox}[
    colback=backcolour,
    colframe=framecolour,
    title=\textbf{Context Window (Truncated)},
    coltitle=black!80,
    fonttitle=\sffamily\small,
    sharp corners,
    boxrule=0.8pt,
    left=4pt, right=4pt, top=2pt, bottom=2pt
]
\begin{lstlisting}[basicstyle=\ttfamily\small]
Keim, SC (2004) "Risk perceptions and safety compliance of workers employed in agriculture, forestry, and fishing." Journal of Safety Research 35(2), 23@\target{7}@-244.
\end{lstlisting}
\end{tcolorbox}

\begin{center}
    \begin{minipage}{\linewidth}
        \centering
        \small
        \setlength{\tabcolsep}{14pt} 
        \captionof{table}{Top-10 Prediction Distribution for Instance 4 ($k=5$).} 
        \label{tab:instance_k5_prob}
        \vspace{-0.3em}
        
        \begin{tabular}{ccc@{\hskip 30pt}ccc} 
        \toprule
        \textbf{Rank} & \textbf{Token} & \textbf{Prob.} & 
        \textbf{Rank} & \textbf{Token} & \textbf{Prob.} \\
        \midrule
        1 & \texttt{5} & 0.198 & 6  & \texttt{-} & 0.003 \\
        2 & \texttt{3} & 0.198 & 7  & \texttt{4} & 0.001 \\
        3 & \texttt{7} & 0.198 & 8  & \texttt{6} & 0.001 \\
        4 & \texttt{1} & 0.198 & 9  & \texttt{0} & 0.001 \\
        5 & \texttt{9} & 0.198 & 10 & \texttt{2} & 0.001 \\
        \bottomrule
        \end{tabular}
    \end{minipage}
\end{center}
\vspace{0.5em}

\paragraph{Instance 5: $k=6$}
\noindent
\textbf{Target Token:} \texttt{4}
\hspace{2em}
\textbf{$N_1$:} $6.021$
\hspace{2em}
\textbf{$R$:} $0.9972$
\begin{tcolorbox}[
    colback=backcolour,
    colframe=framecolour,
    title=\textbf{Context Window (Truncated)},
    coltitle=black!80,
    fonttitle=\sffamily\small,
    sharp corners,
    boxrule=0.8pt,
    left=4pt, right=4pt, top=2pt, bottom=2pt
]
\begin{lstlisting}[basicstyle=\ttfamily\small]
A. G., & Tuuli, M. G. (2016). Maternal marijuana use and adverse neonatal outcomes. Obstetrics & Gynecology, 128(@\target{4}@), 713-723.
\end{lstlisting}
\end{tcolorbox}

\begin{center}
    \begin{minipage}{\linewidth}
        \centering
        \small
        \setlength{\tabcolsep}{14pt} 
        \captionof{table}{Top-10 Prediction Distribution for Instance 5 ($k=6$).} 
        \label{tab:instance_k6_prob}
        \vspace{-0.3em}
        
        \begin{tabular}{ccc@{\hskip 30pt}ccc} 
        \toprule
        \textbf{Rank} & \textbf{Token} & \textbf{Prob.} & 
        \textbf{Rank} & \textbf{Token} & \textbf{Prob.} \\
        \midrule
        1 & \texttt{2} & 0.167 & 6  & \texttt{3} & 0.167 \\
        2 & \texttt{1} & 0.167 & 7  & \texttt{7} & 0.000 \\
        3 & \texttt{6} & 0.167 & 8  & \texttt{9} & 0.000 \\
        4 & \texttt{5} & 0.167 & 9  & \texttt{8} & 0.000 \\
        5 & \texttt{4} & 0.167 & 10 & \texttt{0} & 0.000 \\
        \bottomrule
        \end{tabular}
    \end{minipage}
\end{center}
\vspace{0.5em}

\paragraph{Instance 6: $k=7$}

\noindent
\textbf{Target Token:} \texttt{3}
\hspace{2em}
\textbf{$N_1$:} $7.073$
\hspace{2em}
\textbf{$R$:} $0.9722$
\begin{tcolorbox}[
    colback=backcolour,
    colframe=framecolour,
    title=\textbf{Context Window (Truncated)},
    coltitle=black!80,
    fonttitle=\sffamily\small,
    sharp corners,
    boxrule=0.8pt,
    left=4pt, right=4pt, top=2pt, bottom=2pt
]
\begin{lstlisting}[basicstyle=\ttfamily\small]
According to a 2021 Gallup poll, 54% of Americans support the death penalty for murder, while 4@\target{3}@% oppose it.
\end{lstlisting}
\end{tcolorbox}

\begin{center}
    \begin{minipage}{\linewidth}
        \centering
        \small
        \setlength{\tabcolsep}{14pt} 
        \captionof{table}{Top-10 Prediction Distribution for Instance 6 ($k=7$).} 
        \label{tab:instance_k7_prob}
        \vspace{-0.3em}
        
        \begin{tabular}{ccc@{\hskip 30pt}ccc} 
        \toprule
        \textbf{Rank} & \textbf{Token} & \textbf{Prob.} & 
        \textbf{Rank} & \textbf{Token} & \textbf{Prob.} \\
        \midrule
        1 & \texttt{2} & 0.167 & 6  & \texttt{4} & 0.131 \\
        2 & \texttt{1} & 0.167 & 7  & \texttt{5} & 0.102 \\
        3 & \texttt{6} & 0.167 & 8  & \texttt{7} & 0.003 \\
        4 & \texttt{3} & 0.131 & 9  & \texttt{8} & 0.001 \\
        5 & \texttt{0} & 0.131 & 10 & \texttt{9} & 0.000 \\
        \bottomrule
        \end{tabular}
    \end{minipage}
\end{center}
\vspace{0.5em}

\paragraph{Instance 7: $k=8$}
\noindent
\textbf{Target Token:} \texttt{4}
\hspace{2em}
\textbf{$N_1$:} $8.285$
\hspace{2em}
\textbf{$R$:} $0.9733$
\begin{tcolorbox}[
    colback=backcolour,
    colframe=framecolour,
    title=\textbf{Context Window (Truncated)},
    coltitle=black!80,
    fonttitle=\sffamily\small,
    sharp corners,
    boxrule=0.8pt,
    left=4pt, right=4pt, top=2pt, bottom=2pt
]
\begin{lstlisting}[basicstyle=\ttfamily\small]
'Authorization' => 'Basic NTZmZGJmN2EtODUxYy00M2RiLTk@\target{4}@YWUtYTBhZmEzYzFjZGRi'
\end{lstlisting}
\end{tcolorbox}

\begin{center}
    \begin{minipage}{\linewidth}
        \centering
        \small
        \setlength{\tabcolsep}{14pt} 
        \captionof{table}{Top-10 Prediction Distribution for Instance 7 ($k=8$).} 
        \label{tab:instance_k8_prob}
        \vspace{-0.3em}
        
        \begin{tabular}{ccc@{\hskip 30pt}ccc} 
        \toprule
        \textbf{Rank} & \textbf{Token} & \textbf{Prob.} & 
        \textbf{Rank} & \textbf{Token} & \textbf{Prob.} \\
        \midrule
        1 & \texttt{x} & 0.159 & 6  & \texttt{2} & 0.110 \\
        2 & \texttt{1} & 0.125 & 7  & \texttt{4} & 0.110 \\
        3 & \texttt{z} & 0.125 & 8  & \texttt{5} & 0.110 \\
        4 & \texttt{0} & 0.125 & 9  & \texttt{zM} & 0.011 \\
        5 & \texttt{3} & 0.125 & 10 & \texttt{...} & 0.000 \\
        \bottomrule
        \end{tabular}
    \end{minipage}
\end{center}
\vspace{0.5em}

\paragraph{Instance 8: $k=9$}

\noindent
\textbf{Target Token:} \texttt{6}
\hspace{2em}
\textbf{$N_1$:} $9.033$
\hspace{2em}
\textbf{$R$:} $0.9974$
\begin{tcolorbox}[
    colback=backcolour,
    colframe=framecolour,
    title=\textbf{Context Window (Truncated)},
    coltitle=black!80,
    fonttitle=\sffamily\small,
    sharp corners,
    boxrule=0.8pt,
    left=4pt, right=4pt, top=2pt, bottom=2pt
]
\begin{lstlisting}[basicstyle=\ttfamily\small]
Link: <https://www.inc.com/guides/2010/0@\target{6}@/8-entrepreneurial-myths.html>
\end{lstlisting}
\end{tcolorbox}

\begin{center}
    \begin{minipage}{\linewidth}
        \centering
        \small
        \setlength{\tabcolsep}{14pt} 
        \captionof{table}{Top-10 Prediction Distribution for Instance 8 ($k=9$).} 
        \label{tab:instance_k9_prob}
        \vspace{-0.3em}
        
        \begin{tabular}{ccc@{\hskip 30pt}ccc} 
        \toprule
        \textbf{Rank} & \textbf{Token} & \textbf{Prob.} & 
        \textbf{Rank} & \textbf{Token} & \textbf{Prob.} \\
        \midrule
        1 & \texttt{4} & 0.111 & 6  & \texttt{7} & 0.111 \\
        2 & \texttt{3} & 0.111 & 7  & \texttt{5} & 0.111 \\
        3 & \texttt{1} & 0.111 & 8  & \texttt{2} & 0.111 \\
        4 & \texttt{6} & 0.111 & 9  & \texttt{9} & 0.111 \\
        5 & \texttt{8} & 0.111 & 10 & \texttt{0} & 0.000 \\
        \bottomrule
        \end{tabular}
    \end{minipage}
\end{center}
\vspace{0.5em}

\paragraph{Instance 9: $k=10$}
\noindent
\textbf{Target Token:} \texttt{0}
\hspace{2em}
\textbf{$N_1$:} $10.001$
\hspace{2em}
\textbf{$R$:} $0.9999$
\begin{tcolorbox}[
    colback=backcolour,
    colframe=framecolour,
    title=\textbf{Context Window (Truncated)},
    coltitle=black!80,
    fonttitle=\sffamily\small,
    sharp corners,
    boxrule=0.8pt,
    left=4pt, right=4pt, top=2pt, bottom=2pt
]
\begin{lstlisting}[basicstyle=\ttfamily\small]
| Household Type | Number of Households | Percentage |
| --- | --- | --- |
| Single-person households | 1,598,8@\target{0}@6 | 54.0% |
\end{lstlisting}
\end{tcolorbox}

\begin{center}
    \begin{minipage}{\linewidth}
        \centering
        \small
        \setlength{\tabcolsep}{14pt} 
        \captionof{table}{Top-10 Prediction Distribution for Instance 9 ($k=10$).} 
        \label{tab:instance_k10_prob}
        \vspace{-0.3em}
        
        \begin{tabular}{ccc@{\hskip 30pt}ccc} 
        \toprule
        \textbf{Rank} & \textbf{Token} & \textbf{Prob.} & 
        \textbf{Rank} & \textbf{Token} & \textbf{Prob.} \\
        \midrule
        1 & \texttt{3} & 0.100 & 6  & \texttt{6} & 0.100 \\
        2 & \texttt{2} & 0.100 & 7  & \texttt{0} & 0.100 \\
        3 & \texttt{4} & 0.100 & 8  & \texttt{1} & 0.100 \\
        4 & \texttt{5} & 0.100 & 9  & \texttt{9} & 0.100 \\
        5 & \texttt{7} & 0.100 & 10 & \texttt{8} & 0.100 \\
        \bottomrule
        \end{tabular}
    \end{minipage}
\end{center}
\vspace{0.5em}

\paragraph{Instance 10: $k=12$}
\noindent
\textbf{Target Token:} \texttt{9}
\hspace{2em}
\textbf{$N_1$:} $11.998$
\hspace{2em}
\textbf{$R$:} $0.9981$
\begin{tcolorbox}[
    colback=backcolour,
    colframe=framecolour,
    title=\textbf{Context Window (Truncated)},
    coltitle=black!80,
    fonttitle=\sffamily\small,
    sharp corners,
    boxrule=0.8pt,
    left=4pt, right=4pt, top=2pt, bottom=2pt
]
\begin{lstlisting}[basicstyle=\ttfamily\small]
i.pinimg.com/originals/54/ba/82/54ba8283ca60486b3acd27528d7faec@\target{9}@.jpg) | Set up a dedicated space with simple building blocks ...
\end{lstlisting}
\end{tcolorbox}

\begin{center}
    \begin{minipage}{\linewidth}
        \centering
        \small
        \setlength{\tabcolsep}{14pt} 
        \captionof{table}{Top-20 Prediction Distribution for Instance 10 ($k=12$).} 
        \label{tab:instance_k12_prob}
        \vspace{-0.3em}
        
        \begin{tabular}{ccc@{\hskip 30pt}ccc} 
        \toprule
        \textbf{Rank} & \textbf{Token} & \textbf{Prob.} & 
        \textbf{Rank} & \textbf{Token} & \textbf{Prob.} \\
        \midrule
        1  & \texttt{b} & 0.092 & 11 & \texttt{2} & 0.082 \\
        2  & \texttt{f} & 0.092 & 12 & \texttt{3} & 0.082 \\
        3  & \texttt{7} & 0.082 & 13 & \texttt{d} & 0.000 \\
        4  & \texttt{6} & 0.082 & 14 & \texttt{<eot>} & 0.000 \\
        5  & \texttt{1} & 0.082 & 15 & \texttt{ac} & 0.000 \\
        6  & \texttt{0} & 0.082 & 16 & \texttt{g} & 0.000 \\
        7  & \texttt{4} & 0.082 & 17 & \texttt{ab} & 0.000 \\
        8  & \texttt{5} & 0.082 & 18 & \texttt{fa} & 0.000 \\
        9  & \texttt{8} & 0.082 & 19 & \texttt{=} & 0.000 \\
        10 & \texttt{9} & 0.082 & 20 & \texttt{ad} & 0.000 \\
        \bottomrule
        \end{tabular}
    \end{minipage}
\end{center}
\vspace{0.5em}

\paragraph{Instance 11: $k=15$}
\noindent
\textbf{Target Token:} \texttt{a}
\hspace{2em}
\textbf{$N_1$:} $14.966$
\hspace{2em}
\textbf{$R$:} $0.9969$
\begin{tcolorbox}[
    colback=backcolour,
    colframe=framecolour,
    title=\textbf{Context Window (Truncated)},
    coltitle=black!80,
    fonttitle=\sffamily\small,
    sharp corners,
    boxrule=0.8pt,
    left=4pt, right=4pt, top=2pt, bottom=2pt
]
\begin{lstlisting}[basicstyle=\ttfamily\small]
| STEM Corner | [Image](https://i.pinimg.com/originals/54/b@\target{a}@/82/54ba8283ca60486b3acd27528d7faec9.jpg) | Set up a dedicated space ...
\end{lstlisting}
\end{tcolorbox}

\begin{center}
    \begin{minipage}{\linewidth}
        \centering
        \small
        \setlength{\tabcolsep}{14pt} 
        \captionof{table}{Top-20 Prediction Distribution for Instance 11 ($k=15$).} 
        \label{tab:instance_k15_prob}
        \vspace{-0.3em}
        
        \begin{tabular}{ccc@{\hskip 30pt}ccc} 
        \toprule
        \textbf{Rank} & \textbf{Token} & \textbf{Prob.} & 
        \textbf{Rank} & \textbf{Token} & \textbf{Prob.} \\
        \midrule
        1  & \texttt{4}  & 0.070 & 11 & \texttt{8}  & 0.062 \\
        2  & \texttt{5}  & 0.070 & 12 & \texttt{3}  & 0.062 \\
        3  & \texttt{1}  & 0.070 & 13 & \texttt{7}  & 0.062 \\
        4  & \texttt{9}  & 0.070 & 14 & \texttt{6}  & 0.062 \\
        5  & \texttt{b}  & 0.070 & 15 & \texttt{f}  & 0.055 \\
        6  & \texttt{a}  & 0.070 & 16 & \texttt{e}  & 0.000 \\
        7  & \texttt{d}  & 0.070 & 17 & \texttt{ad} & 0.000 \\
        8  & \texttt{0}  & 0.070 & 18 & \texttt{ab} & 0.000 \\
        9  & \texttt{c}  & 0.070 & 19 & \texttt{fa} & 0.000 \\
        10 & \texttt{2}  & 0.062 & 20 & \texttt{da} & 0.000 \\
        \bottomrule
        \end{tabular}
    \end{minipage}
\end{center}
\vspace{0.5em}

\section{Gradient of the Locally Normalized Preservation Loss}
\label{app:gradient_derivation}

This appendix supplies the derivations behind Section~\ref{sec:gradient_view}. Each subsection establishes one claim made there, in the order the main text uses them:

\begin{itemize}
\itemsep2pt
\item Appendix~\ref{app:grad_normalization} derives Eqs.~\eqref{eq:normalized_gradient_cases}--\eqref{eq:unnorm_gradient_cases}.
\item Appendix~\ref{app:grad_target} derives Eq.~\eqref{eq:target_cap} and Table~\ref{tab:target_cap}, showing that only the complete design leaves the target logit unconstrained.
\end{itemize}

\subsection{Normalized and Unnormalized Preservation Gradients}
\label{app:grad_normalization}

Fix a position $t$ and drop the subscript for readability. Let $p$ and $q$ denote the original next-token distributions of the trainable and base models, respectively, and let $z$ denote the trainable logits. Define
\[
P=\sum_{v\in\mathcal{A}}p(v),\qquad
Q=\sum_{v\in\mathcal{A}}q(v),\qquad
\hat{p}(v)=\frac{p(v)}{P},\qquad
\hat{q}(v)=\frac{q(v)}{Q}.
\]
We refer to $p(v)$ and $q(v)$ as the \emph{original probabilities}, and to $\hat{p}(v)$ and $\hat{q}(v)$ as the \emph{locally normalized probabilities}. The preservation set $\mathcal{A}$ and the base distribution $q$ are fixed during training; for nonempty $\mathcal{A}$, both $P$ and $Q$ are strictly positive.

\paragraph{Normalized gradient.}
For $v\in\mathcal{A}$, local normalization cancels the global softmax normalizer:
\[
\hat{p}(v)
=\frac{\exp z(v)}{\sum_{u\in\mathcal{A}}\exp z(u)},
\qquad
\frac{\partial\ln\hat{p}(v)}{\partial z(w)}
=\mathbf{1}[v=w]-\hat{p}(w)\mathbf{1}[w\in\mathcal{A}].
\]
Since $\mathcal{L}_{\mathrm{LP}}=-\sum_{v\in\mathcal{A}}\hat{q}(v)\ln\hat{p}(v)+\mathrm{const}$ and $\sum_{v\in\mathcal{A}}\hat{q}(v)=1$,
\begin{equation}
\label{eq:lp_gradient}
\frac{\partial\mathcal{L}_{\mathrm{LP}}}{\partial z(w)}
=
\begin{cases}
\hat{p}(w)-\hat{q}(w)
=\dfrac{p(w)}{P}-\dfrac{q(w)}{Q}, & w\in\mathcal{A},\\
0, & w\notin\mathcal{A}.
\end{cases}
\end{equation}

\paragraph{Unnormalized decomposition.}
For $\widetilde{\mathcal{L}}=\sum_{v\in\mathcal{A}}q(v)\ln\frac{q(v)}{p(v)}$, substituting $q(v)=Q\hat{q}(v)$ and $p(v)=P\hat{p}(v)$ gives
\begin{equation}
\begin{aligned}
\widetilde{\mathcal{L}}
&=\sum_{v\in\mathcal{A}}Q\hat{q}(v)
\ln\frac{Q\hat{q}(v)}{P\hat{p}(v)}\\
&=Q\,D_{\mathrm{KL}}\!\left(\hat{q}\,\|\,\hat{p}\right)
+Q\ln\frac{Q}{P}
=Q\mathcal{L}_{\mathrm{LP}}+Q\ln\frac{Q}{P}.
\end{aligned}
\end{equation}

\paragraph{Unnormalized gradient.}
Using $\partial\ln p(v)/\partial z(w)=\mathbf{1}[v=w]-p(w)$,
\begin{equation}
\label{eq:unnorm_grad}
\begin{aligned}
\frac{\partial\widetilde{\mathcal{L}}}{\partial z(w)}
&=-\sum_{v\in\mathcal{A}}q(v)
\left(\mathbf{1}[v=w]-p(w)\right)\\
&=-q(w)\mathbf{1}[w\in\mathcal{A}]+Qp(w)\\
&=
\begin{cases}
Qp(w)-q(w), & w\in\mathcal{A},\\
Qp(w), & w\notin\mathcal{A}.
\end{cases}
\end{aligned}
\end{equation}

\subsection{Stationary Target Probability Under the Four Designs}
\label{app:grad_target}

This section compares the four combinations of target-token removal and local normalization by deriving the stationary target probability under each design. We use $\mathcal{L}_{\mathrm{pres}}$ to denote the preservation term of the design under consideration. In every case,
\[
\frac{\partial \mathcal{L}_{\mathrm{CE}}}{\partial z(y_t)}
=p(y_t)-1.
\]
We add the corresponding preservation gradient and solve the logit-level stationarity condition for $p^{\star}(y_t)$.

\emph{(i) Remove, normalized.} Here $y_t\notin\mathcal{A}$, so Eq.~\eqref{eq:lp_gradient} gives $\partial\mathcal{L}_{\mathrm{pres}}/\partial z(y_t)=0$. The stationarity condition reduces to $p(y_t)-1=0$, and hence $p^{\star}(y_t)=1$ for every $\mu$.

\emph{(ii) Retain, normalized.} Now $\mathcal{A}=S$ and $y_t\in S$. Writing $\rho=p(S)$ gives $\hat{p}(y_t)=p(y_t)/\rho$, so stationarity requires
\[
p(y_t)-1+\mu\left(\frac{p(y_t)}{\rho}-\hat{q}(y_t)\right)=0.
\]
Solving for the target probability yields
\[
p_{\mathrm{pres}}^{\star}(y_t)
=\frac{1+\mu\hat{q}(y_t)}{1+\mu/\rho}
\]
Since $\rho\le1$,
\[
p_{\mathrm{pres}}^{\star}(y_t)
\le\frac{1+\mu\hat{q}(y_t)}{1+\mu}
\le1.
\]
This proves Eq.~\eqref{eq:target_cap}. The final inequality is strict whenever $\hat{q}(y_t)<1$; the exact stationary value equals $1$ only in the degenerate case $\rho=\hat{q}(y_t)=1$. At a plateau-$k$ position with $k>1$, $\hat{q}(y_t)\approx1/k<1$, so retaining the target strictly caps its stationary probability.

\emph{(iii) Remove, unnormalized.} Although $y_t\notin\mathcal{A}$, Eq.~\eqref{eq:unnorm_grad} gives $\partial\widetilde{\mathcal{L}}/\partial z(y_t)=Qp(y_t)>0$, where $Q=q(\mathcal{A})$. Therefore
\[
p(y_t)-1+\mu Qp(y_t)=0,
\qquad
p^{\star}(y_t)=\frac{1}{1+\mu Q}<1.
\]
Removing the target is insufficient: the unnormalized objective suppresses all tokens outside $\mathcal{A}$.

\emph{(iv) Retain, unnormalized.} With $\mathcal{A}=S$, Eq.~\eqref{eq:unnorm_grad} gives
\[
\frac{\partial\widetilde{\mathcal{L}}}{\partial z(y_t)}
=-q(y_t)+q(S)p(y_t),
\qquad
p^{\star}(y_t)
=\frac{1+\mu q(y_t)}{1+\mu q(S)}.
\]
This is strictly below $1$ whenever $S$ contains nonzero base-model mass beyond the target.

Only case (i) guarantees the unmodified cross-entropy fixed point for every nonempty preservation set. Each incomplete design caps the target probability whenever the set contains nonzero non-target mass. The value $p^{\star}(y_t)=1$ expresses compatibility with the cross-entropy objective, not a guarantee of deterministic predictions after training. It means that the preservation term does not attenuate the direct supervised signal at the target logit. Because both losses propagate through shared parameters, indirect interactions may remain.

\begin{table}[t]
\centering
\small
\renewcommand{\arraystretch}{1.35}
\caption{
\textbf{Stationary target probability under each design.}
Only the complete design---target-token removal with local normalization---preserves the cross-entropy fixed point $p_t^\star(y_t)=1$; each incomplete variant caps the target probability below one. Exemplary values are computed with $\mu = 1$ under an idealized plateau-$3$ state ($q_t(y_t)=1/3$, $\rho_t = 1$, and $q_t(\mathcal{A}_t) = 2/3$) to illustrate the relative capping effect.
}
\label{tab:target_cap}
\begin{tabular}{@{}cc l l c@{}}
\toprule
\makecell{\textbf{Remove} $\boldsymbol{y_t}$} & \makecell{\textbf{Normalized}\\\textbf{KL}}
& $\partial \mathcal{L}_{\mathrm{pres}}^t / \partial z_t(y_t)$
& $p_t^{\star}(y_t)$ & \textbf{Value} \\
\midrule
\rowcolor{BAShade}
Yes & Yes & $0$ & $1$ & $1.00$ \\
No  & Yes & $\hat{p}_t(y_t) - \hat{q}_t(y_t)$ & $\dfrac{1+\mu\,\hat{q}_t(y_t)}{1+\mu/\rho_t}$ & $0.67$ \\
Yes & No  & $q_t(\mathcal{A}_t)\,p_t(y_t)$ & $\dfrac{1}{1+\mu\,q_t(\mathcal{A}_t)}$ & $0.60$ \\
No  & No  & $-q_t(y_t) + q_t(S_t)\,p_t(y_t)$ & $\dfrac{1+\mu\,q_t(y_t)}{1+\mu\,q_t(S_t)}$ & $0.67$ \\
\bottomrule
\end{tabular}
\end{table}

\section{LP-SFT Algorithm Pseudocode}
\label{app:algorithm}

Algorithm~\ref{alg:lpsft} presents the complete pseudocode for LP-SFT, detailing the two-stage implementation: an offline stage to cache base model predictions, and an online stage for efficient fine-tuning using only the trainable model.

\begin{algorithm}[ht]
\caption{LP-SFT Training}
\label{alg:lpsft}
\small
\begin{algorithmic}[1]
\REQUIRE Base model $\theta_0$, training dataset $\mathcal{D}$, support size $K_{\max}$, weight $\mu$
\STATE \textbf{// Stage 1: Offline Base Precomputation}
\FOR{each sample $(\mathbf{x}, \mathbf{y}) \in \mathcal{D}$}
  \STATE Compute $q_t = p_{\theta_0}(\cdot \mid \mathbf{x}, \mathbf{y}_{<t})$ and cache top-$K_{\max}$ candidate logits for all valid $t$
\ENDFOR
\STATE \textbf{// Stage 2: Local-Preserving Supervised Fine-Tuning}
\FOR{each mini-batch from $\mathcal{D}$}
  \STATE $\mathcal{L} \gets 0$
  \FOR{each target position $t$}
    \STATE Load cached base logits, set $S_t=C_{K_{\max}}(q_t)$, and construct $\mathcal{A}_t$ via Eq.~\eqref{eq:action_set}
    \STATE $\mathcal{L}_{\mathrm{LP}}^t \gets 0$
    \IF{$\mathcal{A}_t \neq \emptyset$}
      \STATE $\hat{q}_t, \hat{p}_t \gets \mathrm{Normalize}(q_t|_{\mathcal{A}_t}), \mathrm{Normalize}(p_t|_{\mathcal{A}_t})$
      \STATE $\mathcal{L}_{\mathrm{LP}}^t \gets \sum_{v \in \mathcal{A}_t} \hat{q}_t(v) \ln \frac{\hat{q}_t(v)}{\hat{p}_t(v)}$
    \ENDIF
    \STATE $\mathcal{L} \gets \mathcal{L} - \ln p_t(y_t) + \mu \cdot \mathcal{L}_{\mathrm{LP}}^t$
  \ENDFOR
  \STATE Update parameters $\theta$ using $\nabla_\theta \mathcal{L}$
\ENDFOR
\end{algorithmic}
\end{algorithm}

\section{Training and Evaluation Details} 
\label{app:training_eval_details} 

\subsection{Training Details} 
\label{app:training_details} 

All models are fine-tuned using the HuggingFace Transformers library with DeepSpeed for memory-efficient distributed training. We use ZeRO Stage-2 by default, and ZeRO Stage-3 for ASFT, which keeps a frozen reference model in memory during training. We use the AdamW optimizer with a learning rate of $2\times 10^{-5}$, a cosine learning-rate schedule, and a warmup ratio of $0.03$. The weight decay is set to $0.0$, and we use $\beta_2=0.95$ for AdamW. All models are trained for 3 epochs in bfloat16 precision with gradient checkpointing enabled. The maximum sequence length is set to 2048 tokens, and the random seed is fixed to 1234 for reproducibility. We use an effective batch size of 128 across all experiments, adjusting the per-device micro-batch size and gradient accumulation steps according to model scale and method-specific memory usage (e.g., a smaller micro-batch for Qwen3-14B and for ASFT). All experiments are conducted on nodes equipped with 8$\times$ NVIDIA A100 (80GB) GPUs.

\paragraph{Baselines.}
For the GEM baseline, we set the temperature parameter to $\beta=0.7$. For EAFT, we use $\alpha=1.0$ and top-$k=20$. For ASFT, we set the weight of the full-vocabulary KL loss to $0.05$. These hyperparameter choices follow the recommended settings in the respective original papers.

\subsection{Evaluation Details} 
\label{app:evaluation_details} 

\paragraph{General evaluation protocol.}
All SFT models are evaluated using the chat prompt template of their respective model family (ChatML for Qwen3 and the Llama-3.1 instruct format for Llama-3.1-8B), with thinking mode disabled for Qwen3 models (\texttt{enable\_thinking=False}) to ensure a fair comparison under standard non-thinking generation. For base models, we use the corresponding base-model prompting format when evaluating generation tasks, without applying instruction-tuned chat-specific behavior. MMLU is the sole exception, where all models are evaluated with the base non-chat prompt format to avoid formatting artifacts in log-likelihood scoring. We employ vLLM as the inference backend across all benchmarks to accelerate evaluation.

\paragraph{Mathematical reasoning.}
For MATH-500, we use temperature $T=0.6$, $n=16$ samples, $\texttt{top\_p}=1.0$, and a maximum generation length of 4096 tokens ($\texttt{max\_model\_len}=8192$). We use a boxed-answer prompt template, extract final answers from \texttt{\textbackslash boxed\{\}}, and report pass@1 and pass@16. For AIME, we use temperature $T=0.6$, $n=32$ samples, $\texttt{top\_p}=1.0$, a maximum generation length of 8192 tokens, and $\texttt{max\_model\_len}=12288$. The AIME evaluation is conducted on 180 problems collected from AIME competitions from 2021 to 2026, and we report pass@1 and pass@32.

\paragraph{Code generation.} 
HumanEval+ and MBPP+ are evaluated using the EvalPlus framework. We use temperature $T=0.8$ and generate $n=20$ samples per problem. Unless otherwise noted, we report unbiased pass@1 and pass@10 on the plus tests with additional hidden test cases.

\paragraph{General knowledge and reasoning.} 
For general knowledge and reasoning, we evaluate MMLU in the 5-shot setting using log-likelihood scoring with the lm-evaluation-harness framework.

\section{Additional Results}
\label{app:additional_results}

\subsection{Overview}
\label{app:additional_results_overview}

We first report the pretrained reference performance of Qwen3-4B, Qwen3-14B, and Llama-3.1-8B in Table~\ref{tab:app_base_results}, providing the baseline needed to assess capability gains and forgetting after fine-tuning. Tables~\ref{tab:app_magicoder_results} and~\ref{tab:app_numina_results} then present the complete fine-tuning results on Magicoder-OSS-Instruct-75K and NuminaMath-CoT, respectively, complementing the UltraFeedback results in the main text. Finally, Table~\ref{tab:app_qwen25_72b_results} evaluates whether LP-SFT continues to provide a favorable accuracy--preservation trade-off when scaled to Qwen2.5-72B, while Table~\ref{tab:app_qwen25_72b_training_time} compares the corresponding LoRA fine-tuning and cache-construction costs. All evaluations follow the main-text protocol.

\begin{table*}[t]
\centering
\scriptsize
\setlength{\tabcolsep}{5pt}
\renewcommand{\arraystretch}{1.08}
\caption{
\textbf{Base model reference scores.}
Qwen3-14B provides the strongest pretrained reference across the evaluation suite, followed by Qwen3-4B and Llama-3.1-8B.
}
\label{tab:app_base_results}

\begin{tabular}{@{}l cccc cccc c c@{}}
\toprule
\multirow{3}{*}{\textbf{Model}}
& \multicolumn{4}{c}{\textbf{Math}}
& \multicolumn{4}{c}{\textbf{Code}}
& \multicolumn{1}{c}{\textbf{GEN.}}
& \multirow{2}{*}{\textbf{Avg.}} \\
\cmidrule(lr){2-5}
\cmidrule(lr){6-9}
\cmidrule(lr){10-10}
& \multicolumn{2}{c}{\makecell{\textbf{MATH-500}\\pass@1 / pass@16}}
& \multicolumn{2}{c}{\makecell{\textbf{AIME 21--26}\\pass@1 / pass@32}}
& \multicolumn{2}{c}{\makecell{\textbf{MBPP+}\\pass@1 / pass@10}}
& \multicolumn{2}{c}{\makecell{\textbf{HE+}\\pass@1 / pass@10}}
& \makecell{\textbf{MMLU}\\5-shot Acc.}
& \makecell{ \\ All} \\
\midrule
Qwen3-4B-Base   & 34.58 & 80.80 & 5.71 & 23.89 & 59.95 & 82.29 & 67.74 & 90.83 & 73.02 & 57.65 \\
Qwen3-14B-Base  & 69.50 & 89.40 & 7.00 & 26.11 & 69.62 & 87.88 & 77.74 & 93.32 & 80.51 & 66.79 \\
Llama-3.1-8B    & 8.64 & 44.20 & 0.03 & 1.11 & 42.74 & 72.12 & 27.41 & 66.21 & 65.28 & 36.42 \\
\bottomrule
\end{tabular}

\end{table*}

\begin{table*}[ht]
\centering
\scriptsize
\setlength{\tabcolsep}{5pt}
\renewcommand{\arraystretch}{1.08}
\caption{
\textbf{Fine-tuning results on Magicoder-OSS-Instruct-75K.}
LP-SFT achieves the highest overall average for every model family, driven by strong finite-budget performance and consistent preservation of MMLU accuracy.
}
\label{tab:app_magicoder_results}

\begin{tabular}{@{}ll cccc cccc c c@{}}
\toprule
\multirow{3}{*}{\textbf{Model}}
& \multirow{3}{*}{\textbf{Method}}
& \multicolumn{4}{c}{\textbf{Math}}
& \multicolumn{4}{c}{\textbf{Code}}
& \multicolumn{1}{c}{\textbf{GEN.}}
& \multirow{2}{*}{\textbf{Avg.}} \\
\cmidrule(lr){3-6}
\cmidrule(lr){7-10}
\cmidrule(lr){11-11}
&
& \multicolumn{2}{c}{\makecell{\textbf{MATH-500}\\pass@1 / pass@16}}
& \multicolumn{2}{c}{\makecell{\textbf{AIME 21--26}\\pass@1 / pass@32}}
& \multicolumn{2}{c}{\makecell{\textbf{MBPP+}\\pass@1 / pass@10}}
& \multicolumn{2}{c}{\makecell{\textbf{HE+}\\pass@1 / pass@10}}
& \makecell{\textbf{MMLU}\\5-shot Acc.}
& \makecell{ \\ All} \\
\midrule

\multirow{6}{*}{Qwen3-4B}
& CE    & 34.59 & 78.80 & 2.12 & 13.89 & 63.16 & 75.73 & 71.13 & 82.41 & 71.87 & 54.86 \\
& DFT   & 54.55 & 70.80 & 2.48 & 13.89 & 61.69 & 65.77 & 64.27 & 68.53 & 71.61 & 52.62 \\
& EAFT  & 44.51 & 81.40 & 2.76 & 20.00 & 63.36 & 77.26 & \textbf{71.25} & 86.68 & 72.10 & 57.70 \\
& GEM   & 32.31 & 77.80 & 2.31 & 16.67 & 62.71 & 78.93 & 68.87 & 87.42 & 71.76 & 55.42 \\
& ASFT  & 53.94 & 81.40 & 4.24 & 19.44 & \textbf{64.25} & 73.17 & 68.78 & 81.54 & 72.30 & 57.67 \\
\rowcolor{BAShade} \multirow{-6}{*}{\cellcolor{white}Qwen3-4B}
& LP-SFT & \textbf{55.44} & \textbf{84.80} & \textbf{5.19} & \textbf{21.67} & 62.01 & \textbf{81.74} & 70.09 & \textbf{89.97} & \textbf{72.82} & \textbf{60.41} \\
\midrule

\multirow{6}{*}{Qwen3-14B}
& CE    & 47.35 & 82.20 & 4.01 & 20.00 & 69.58 & 79.94 & \textbf{76.59} & 86.87 & 79.90 & 60.72 \\
& DFT   & \textbf{65.34} & 78.80 & 4.81 & 12.22 & 66.92 & 70.95 & 68.93 & 72.04 & 80.01 & 57.78 \\
& EAFT  & 22.86 & 81.60 & 1.77 & 16.67 & 70.98 & 83.35 & 75.37 & 88.29 & 80.12 & 57.89 \\
& GEM   & 52.16 & 86.20 & 2.92 & 16.67 & 68.89 & 84.77 & 74.76 & 88.25 & 79.80 & 61.60 \\
& ASFT  & 63.54 & 83.20 & 5.14 & 21.11 & 70.05 & 78.78 & 76.25 & 84.89 & 80.11 & 62.56 \\
\rowcolor{BAShade} \multirow{-6}{*}{\cellcolor{white}Qwen3-14B}
& LP-SFT & 55.07 & \textbf{86.60} & \textbf{5.54} & \textbf{23.33} & \textbf{71.31} & \textbf{85.40} & 76.22 & \textbf{91.39} & \textbf{80.61} & \textbf{63.94} \\
\midrule

\multirow{6}{*}{Llama-3.1-8B}
& CE    & 7.34 & 33.40 & 0.05 & 1.67 & \textbf{47.63} & 63.11 & \textbf{41.92} & 62.05 & 61.69 & 35.43 \\
& DFT   & 6.01 & 15.80 & 0.00 & 0.00 & 30.70 & 35.98 & 22.74 & 25.81 & 59.44 & 21.83 \\
& EAFT  & 6.94 & 33.60 & 0.02 & 0.56 & 42.01 & 60.73 & 37.41 & 62.01 & 60.60 & 33.76 \\
& GEM   & 6.20 & 34.20 & \textbf{0.14} & \textbf{3.89} & 46.49 & 65.39 & 39.91 & \textbf{64.67} & 61.27 & 35.80 \\
& ASFT  & 6.53 & 27.60 & 0.07 & 2.22 & 43.36 & 57.21 & 31.83 & 48.05 & 60.77 & 30.85 \\
\rowcolor{BAShade} \multirow{-6}{*}{\cellcolor{white}Llama-3.1-8B}
& LP-SFT & \textbf{8.28} & \textbf{38.60} & 0.07 & 2.22 & 45.70 & \textbf{66.78} & 37.62 & 62.67 & \textbf{62.06} & \textbf{36.00} \\
\bottomrule
\end{tabular}

\end{table*}

\begin{table*}[!t]
\centering
\scriptsize
\setlength{\tabcolsep}{5pt}
\renewcommand{\arraystretch}{1.08}
\caption{
\textbf{Fine-tuning results on NuminaMath-CoT.}
LP-SFT achieves the highest overall average for all three model families, combining strong mathematical reasoning with better retention on code and general knowledge.
}
\label{tab:app_numina_results}

\begin{tabular}{@{}ll cccc cccc c c@{}}
\toprule
\multirow{3}{*}{\textbf{Model}}
& \multirow{3}{*}{\textbf{Method}}
& \multicolumn{4}{c}{\textbf{Math}}
& \multicolumn{4}{c}{\textbf{Code}}
& \multicolumn{1}{c}{\textbf{GEN.}}
& \multirow{2}{*}{\textbf{Avg.}} \\
\cmidrule(lr){3-6}
\cmidrule(lr){7-10}
\cmidrule(lr){11-11}
&
& \multicolumn{2}{c}{\makecell{\textbf{MATH-500}\\pass@1 / pass@16}}
& \multicolumn{2}{c}{\makecell{\textbf{AIME 21--26}\\pass@1 / pass@32}}
& \multicolumn{2}{c}{\makecell{\textbf{MBPP+}\\pass@1 / pass@10}}
& \multicolumn{2}{c}{\makecell{\textbf{HE+}\\pass@1 / pass@10}}
& \makecell{\textbf{MMLU}\\5-shot Acc.}
& \makecell{ \\ All} \\
\midrule

\multirow{6}{*}{Qwen3-4B}
& CE    & 57.30 & 86.00 & 4.11 & 21.11 & 60.71 & 78.90 & \textbf{73.11} & 90.05 & 71.89 & 60.35 \\
& DFT   & 40.36 & 68.60 & 2.20 & 11.67 & \textbf{64.72} & 70.44 & 71.01 & 81.23 & 71.36 & 53.51 \\
& EAFT  & 47.61 & 84.60 & 3.92 & 17.78 & 61.18 & 80.74 & 72.71 & \textbf{91.46} & 71.72 & 59.08 \\
& GEM   & 51.92 & 86.00 & 3.00 & 20.56 & 56.87 & 79.00 & 70.03 & 89.93 & 72.30 & 58.85 \\
& ASFT  & \textbf{61.54} & 83.80 & 4.06 & 16.67 & 63.73 & 77.98 & 71.31 & 86.75 & 72.59 & 59.83 \\
\rowcolor{BAShade} \multirow{-6}{*}{\cellcolor{white}Qwen3-4B}
& LP-SFT & 60.71 & \textbf{86.20} & \textbf{5.31} & \textbf{22.78} & 59.76 & \textbf{81.02} & 71.34 & 91.42 & \textbf{72.67} & \textbf{61.25} \\
\midrule

\multirow{6}{*}{Qwen3-14B}
& CE    & 64.50 & \textbf{90.80} & 5.85 & 26.11 & 68.78 & 85.67 & 78.60 & 93.11 & 79.22 & 65.85 \\
& DFT   & 48.30 & 76.60 & 3.07 & 10.56 & \textbf{72.80} & 78.06 & \textbf{79.27} & 84.89 & 79.55 & 59.23 \\
& EAFT  & 58.80 & 88.20 & 4.06 & 25.00 & 69.55 & 86.30 & 79.15 & 92.44 & 79.08 & 64.73 \\
& GEM   & 61.48 & 90.00 & 4.95 & 26.11 & 63.65 & 86.22 & 75.95 & 92.91 & 79.43 & 64.52 \\
& ASFT  & 66.71 & 88.40 & 5.45 & 21.67 & 72.71 & 83.05 & 78.87 & 90.85 & 79.93 & 65.29 \\
\rowcolor{BAShade} \multirow{-6}{*}{\cellcolor{white}Qwen3-14B}
& LP-SFT & \textbf{66.92} & 90.20 & \textbf{7.19} & \textbf{29.44} & 69.58 & \textbf{86.60} & 77.23 & \textbf{94.03} & \textbf{80.34} & \textbf{66.84} \\
\midrule

\multirow{6}{*}{Llama-3.1-8B}
& CE    & \textbf{31.67} & 70.00 & \textbf{0.61} & 10.00 & 30.09 & 60.80 & 24.09 & 58.85 & 60.61 & 38.52 \\
& DFT   & 13.90 & 27.60 & 0.10 & 2.22 & 21.75 & 32.17 & 8.41 & 15.89 & 55.14 & 19.69 \\
& EAFT  & 29.43 & 67.60 & 0.57 & \textbf{12.22} & 31.94 & 63.47 & 22.96 & 57.43 & 60.00 & 38.40 \\
& GEM   & 27.12 & \textbf{70.60} & 0.52 & 10.56 & 25.50 & 59.78 & 20.40 & 52.43 & 60.55 & 36.38 \\
& ASFT  & 26.38 & 60.40 & 0.54 & 7.78 & 31.87 & 57.37 & 22.29 & 48.43 & 61.07 & 35.13 \\
\rowcolor{BAShade} \multirow{-6}{*}{\cellcolor{white}Llama-3.1-8B}
& LP-SFT & 25.42 & 66.20 & 0.31 & 8.33 & \textbf{36.10} & \textbf{65.92} & \textbf{27.10} & \textbf{61.91} & \textbf{62.49} & \textbf{39.31} \\
\bottomrule
\end{tabular}

\end{table*}

\subsection{Scaling to Qwen2.5-72B}
\label{app:qwen25_72b_results}

To examine whether LP-SFT remains effective at a substantially larger scale, we fine-tune Qwen2.5-72B-Base on a 30K subset of UltraFeedback. For a controlled comparison, all methods use rank-64 LoRA on the attention and MLP projections and share the same three-epoch optimization setup with AdamW, a learning rate of $2\times10^{-4}$, and an effective batch size of 128. For ASFT, we use the LoRA-ASFT variant proposed in the original paper~\citep{zhu2025anchored}. LP-SFT uses $\mu=1$ and a cached top-$10$ non-label reference set, avoiding a frozen reference model during training. As shown in Table~\ref{tab:app_qwen25_72b_results}, LP-SFT achieves the best aggregate performance among the fine-tuned methods, leading on MATH-500 and AIME while retaining high pass@$k$ on both code benchmarks. Although ASFT is stronger on code pass@1, LP-SFT provides a better overall balance and slightly exceeds the base model in average performance.

\begin{table*}[t]
\centering
\scriptsize
\setlength{\tabcolsep}{5pt}
\renewcommand{\arraystretch}{1.08}
\caption{
\textbf{Fine-tuning results on Qwen2.5-72B-Base with UltraFeedback-30K.}
LP-SFT attains the best aggregate score among fine-tuned methods, leading on both mathematical benchmarks and on pass@$k$ for both code benchmarks.
}
\label{tab:app_qwen25_72b_results}

\begin{tabular}{@{}ll cccc cccc c c@{}}
\toprule
\multirow{3}{*}{\textbf{Model}}
& \multirow{3}{*}{\textbf{Method}}
& \multicolumn{4}{c}{\textbf{Math}}
& \multicolumn{4}{c}{\textbf{Code}}
& \multicolumn{1}{c}{\textbf{GEN.}}
& \multirow{2}{*}{\textbf{Avg.}} \\
\cmidrule(lr){3-6}
\cmidrule(lr){7-10}
\cmidrule(lr){11-11}
&
& \multicolumn{2}{c}{\makecell{\textbf{MATH-500}\\pass@1 / pass@16}}
& \multicolumn{2}{c}{\makecell{\textbf{AIME 21--26}\\pass@1 / pass@32}}
& \multicolumn{2}{c}{\makecell{\textbf{MBPP+}\\pass@1 / pass@10}}
& \multicolumn{2}{c}{\makecell{\textbf{HE+}\\pass@1 / pass@10}}
& \makecell{\textbf{MMLU}\\5-shot Acc.}
& \makecell{ \\ All} \\
\midrule

\multirow{7}{*}{Qwen2.5-72B}
& Base  & 47.94 & 85.60 & 4.60 & 22.22 & 70.60 & 87.50 & 74.60 & 92.00 & 86.01 & 63.45 \\
\cmidrule(lr){2-12}
& CE    & 41.81 & 79.60 & 1.72 & 17.22 & 48.10 & 78.70 & 50.50 & 85.00 & 83.68 & 54.04 \\
& DFT   & 52.89 & 70.20 & 2.71 & 10.00 & 63.70 & 67.20 & 70.60 & 76.40 & 83.21 & 55.21 \\
& EAFT  & 42.74 & 79.80 & 1.79 & 16.67 & 49.50 & 78.30 & 51.90 & 86.00 & 84.15 & 54.54 \\
& GEM   & 36.21 & 76.40 & 1.25 & 15.00 & 36.20 & 75.00 & 36.90 & 80.90 & 83.34 & 49.02 \\
& ASFT  & 55.55 & 81.40 & 3.23 & 17.22 & \textbf{69.40} & 78.90 & \textbf{73.10} & 85.30 & 85.54 & 61.07 \\
\rowcolor{BAShade} \cellcolor{white}
& LP-SFT & \textbf{55.93} & \textbf{88.40} & \textbf{5.64} & \textbf{25.00} & 66.10 & \textbf{86.10} & 69.60 & \textbf{91.40} & \textbf{85.66} & \textbf{63.76} \\
\bottomrule
\end{tabular}
\end{table*}

\paragraph{Training time.}
Table~\ref{tab:app_qwen25_72b_training_time} reports cost on 8$\times$ A100 (80GB) GPUs, excluding evaluation. LP-SFT matches the fine-tuning cost of CE. Constructing its top-$10$ cache for the 30K training examples takes 11,232 seconds on the same eight GPUs, or 25.0 A100 GPU-hours. This is a one-time preprocessing cost: the cache is reused across epochs, fine-tuning runs, and ablations with the same base model and data.

\begin{table}[h]
\centering
\scriptsize
\setlength{\tabcolsep}{5pt}
\renewcommand{\arraystretch}{1.05}
\caption{\textbf{Qwen2.5-72B LoRA cost on UltraFeedback-30K.}
LP-SFT incurs negligible fine-tuning overhead over CE. Its first-run total includes the one-time reusable top-$10$ cache construction.}
\label{tab:app_qwen25_72b_training_time}
\begin{tabular*}{0.85\linewidth}{@{\extracolsep{\fill}}lccc@{}}
\toprule
\textbf{Method} &
\makecell{\textbf{One-time cache}\\\textbf{GPU-hours}} &
\makecell{\textbf{Fine-tuning}\\\textbf{GPU-hours}} &
\makecell{\textbf{First-run total}\\\textbf{GPU-hours}} \\
\midrule
CE     & ---  & 120.0 & 120.0 \\
DFT    & ---  & 120.7 & 120.7 \\
EAFT   & ---  & 119.8 & 119.8 \\
GEM    & ---  & 118.2 & 118.2 \\
ASFT   & ---  & 238.4 & 238.4 \\
LP-SFT & 25.0 & 120.2 & 145.2 \\
\bottomrule
\end{tabular*}
\end{table}

\section{Additional Ablation and Efficiency Results}
\label{app:ablation_efficiency}

This section analyzes sensitivity to the preservation weight $\mu$, the design principles, and support size $K_t$. All experiments use Qwen3-4B-Base fine-tuned on UltraFeedback.

\subsection{Sensitivity to the Preservation Weight}
\label{app:mu_ablation}

\begin{figure}[ht]
    \centering
    \includegraphics[width=0.45\linewidth]{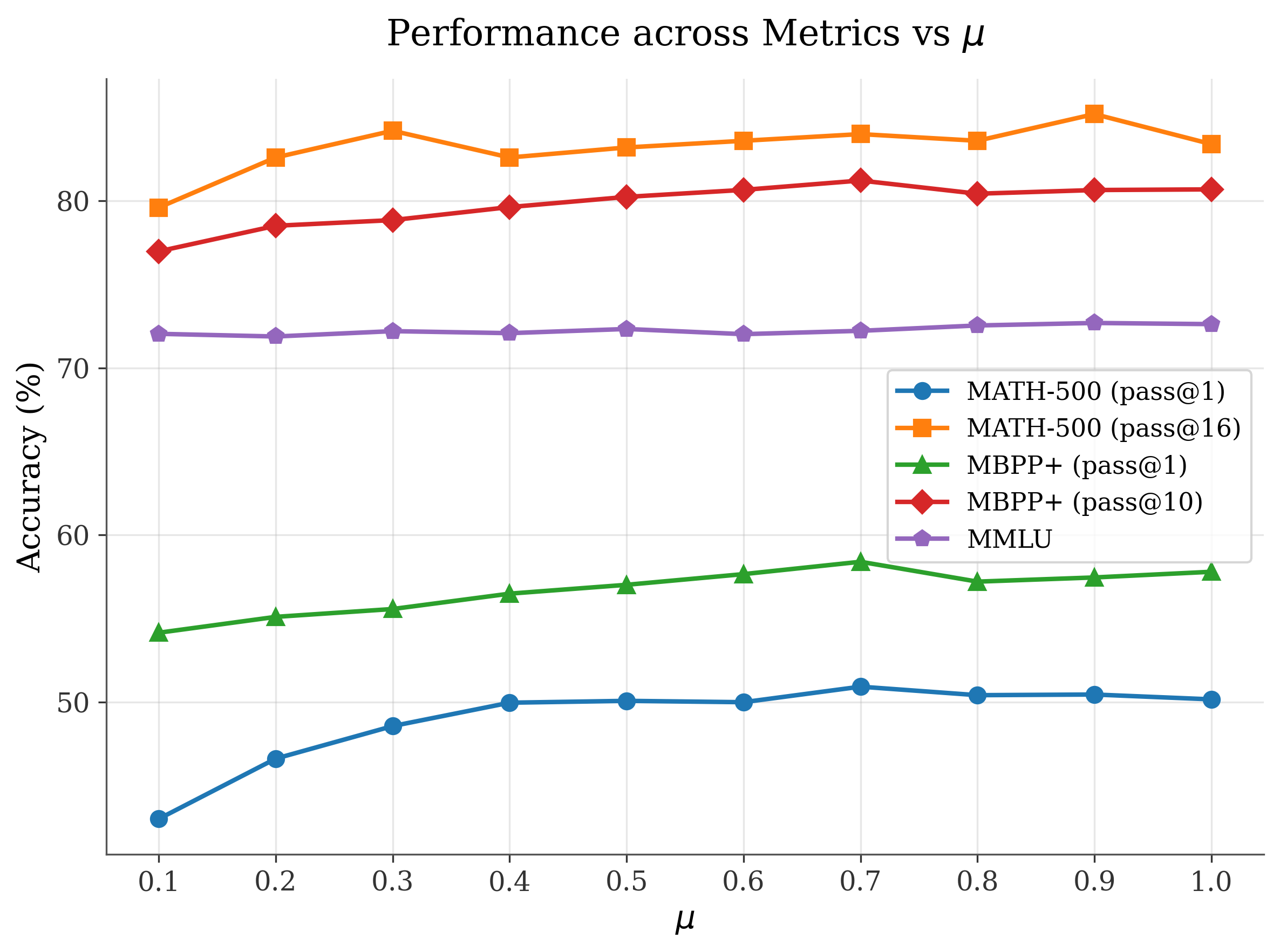}
    \caption{
    \textbf{Sensitivity to the preservation weight $\mu$.}
    Results are obtained with Qwen3-4B fine-tuned on UltraFeedback and evaluated on MATH-500, MBPP+, and MMLU.
    Performance varies only slightly across $\mu \in [0.1,1.0]$, suggesting that LP-SFT is relatively insensitive to the choice of $\mu$.
    }
    \label{fig:mu_ablation}
\end{figure}

Figure~\ref{fig:mu_ablation} studies the effect of the preservation weight $\mu$. Across the tested range, LP-SFT exhibits only mild performance variation on MATH-500, MBPP+, and MMLU, with some metrics showing a slight improvement as $\mu$ increases. This indicates limited sensitivity to $\mu$ within the tested range. We therefore use $\mu=1$ as the default setting in our main experiments.

\subsection{Ablation of the Design Principles}
\label{app:preservation_set_ablation}

Table~\ref{tab:lpsft_ablation_results} evaluates the two design principles on Qwen3-4B-Base trained with UltraFeedback by varying whether the supervised token $y_t$ is removed from the preservation set $\mathcal{A}_t$ and whether locally normalized probabilities are used in the KL loss. Any incomplete configuration falls notably short of the complete objective ($56.77$--$57.49$ vs. $58.79$), confirming that both design principles are necessary to realize the full preservation benefit. This empirical observation aligns with the gradient analysis in Section~\ref{sec:gradient_view} and the target-logit stationary points in Table~\ref{tab:target_cap}. Overall, target-token removal and local normalization determine \emph{whether} preservation helps; we study \emph{how broadly} the constraint is applied next.

\begin{table*}[t]
\centering
\scriptsize
\setlength{\tabcolsep}{4.0pt}
\renewcommand{\arraystretch}{1.08}
\caption{
\textbf{Ablation of LP-SFT design principles on Qwen3-4B trained with UltraFeedback.}
The complete design achieves the highest aggregate score; any incomplete variant (omitting target-token removal, local normalization, or both) underperforms the full objective ($56.77$--$57.49$ vs. $58.79$).
}
\label{tab:lpsft_ablation_results}

\begin{tabular}{@{}cc cccc cccc c c@{}}
\toprule
\multirow{3}{*}{\makecell{\textbf{Remove}\\$\boldsymbol{y_t}$}}
& \multirow{3}{*}{\makecell{\textbf{Normalized}\\\textbf{KL}}}
& \multicolumn{4}{c}{\textbf{Math}}
& \multicolumn{4}{c}{\textbf{Code}}
& \multicolumn{1}{c}{\textbf{GEN.}}
& \multirow{2}{*}{\textbf{Avg.}} \\
\cmidrule(lr){3-6}
\cmidrule(lr){7-10}
\cmidrule(lr){11-11}
&
& \multicolumn{2}{c}{\makecell{\textbf{MATH-500}\\pass@1 / pass@16}}
& \multicolumn{2}{c}{\makecell{\textbf{AIME 21--26}\\pass@1 / pass@32}}
& \multicolumn{2}{c}{\makecell{\textbf{MBPP+}\\pass@1 / pass@10}}
& \multicolumn{2}{c}{\makecell{\textbf{HE+}\\pass@1 / pass@10}}
& \makecell{\textbf{MMLU}\\5-shot Acc.}
& \makecell{All} \\
\midrule

No  & Yes
& 45.99 & 84.20 & 3.52 & 20.00 & 54.30 & 80.04 & 64.88 & 89.22 & 72.35 & 57.17 \\

Yes & No
& 45.71 & \textbf{84.80} & \textbf{3.61} & 20.00 & 51.57 & 80.08 & 64.21 & 88.80 & 72.14 & 56.77 \\

No  & No
& 48.25 & 84.00 & \textbf{3.61} & \textbf{20.56} & 54.17 & 80.28 & 64.73 & 89.59 & 72.20 & 57.49 \\

\midrule

\rowcolor{BAShade}
Yes & Yes
& \textbf{50.15} & 83.40 & 3.39 & 19.44 & \textbf{57.79} & \textbf{80.69} & \textbf{69.30} & \textbf{92.29} & \textbf{72.62} & \textbf{58.79} \\

\bottomrule
\end{tabular}
\end{table*}

\begin{table*}[t]
    \centering
    \scriptsize
    \setlength{\tabcolsep}{3pt}
    \renewcommand{\arraystretch}{1.08}
    \caption{
    \textbf{Ablation on support size $K_t$ (Qwen3-4B-Base, UltraFeedback, $\mu=1.0$).}
    Fixed top-$10$ preservation comes within 0.22 average points of full-vocabulary preservation while using only an offline cache rather than an in-memory reference model.
    }
    \label{tab:uf_support_schedule}
    
    \begin{tabular}{@{}l l cccc cccc c c@{}}
    \toprule
    \multirow{3}{*}{\textbf{Support schedule}}
    & \multirow{3}{*}{\makecell[l]{\textbf{Base}\\\textbf{access}}}
    & \multicolumn{4}{c}{\textbf{Math}}
    & \multicolumn{4}{c}{\textbf{Code}}
    & \multicolumn{1}{c}{\textbf{GEN.}}
    & \multirow{2}{*}{\textbf{Avg.}} \\
    \cmidrule(lr){3-6}
    \cmidrule(lr){7-10}
    \cmidrule(lr){11-11}
    &
    & \multicolumn{2}{c}{\makecell{\textbf{MATH-500}\\pass@1 / pass@16}}
    & \multicolumn{2}{c}{\makecell{\textbf{AIME 21--26}\\pass@1 / pass@32}}
    & \multicolumn{2}{c}{\makecell{\textbf{MBPP+}\\pass@1 / pass@10}}
    & \multicolumn{2}{c}{\makecell{\textbf{HE+}\\pass@1 / pass@10}}
    & \makecell{\textbf{MMLU}\\5-shot Acc.}
    & \makecell{All} \\
    \midrule
    
    \makecell[l]{$N_1$-adaptive\\$K_t=\lceil N_1(q_t)\rceil$} & Cache
    & 48.77 & 83.20 & 2.74 & \textbf{19.44}
    & 55.69 & 79.17 & 68.93 & 89.41
    & 72.28 & 57.74 \\
    
    \rowcolor{BAShade}
    \makecell[l]{Fixed top-$10$\\(default)} & Cache
    & 50.15 & 83.40 & \textbf{3.39} & \textbf{19.44}
    & 57.79 & 80.69 & \textbf{69.30} & \textbf{92.29}
    & 72.62 & 58.79 \\
    
    \makecell[l]{Full non-label\\vocabulary} & Ref.\ model
    & \textbf{52.35} & \textbf{85.00} & \textbf{3.39} & \textbf{19.44}
    & \textbf{58.57} & \textbf{81.60} & 67.96 & 90.10
    & \textbf{72.67} & \textbf{59.01} \\
    
    \bottomrule
    \end{tabular}
\end{table*}

\subsection{Ablation of the Support Size \texorpdfstring{$K_t$}{K-t}}
\label{app:preservation_Kt}

Having fixed the two design principles, we next vary only the breadth of the candidate set $S_t=C_{K_t}(q_t)$ in Eq.~\eqref{eq:action_set}. Table~\ref{tab:uf_support_schedule} compares three choices of $K_t$ under an otherwise identical setup (Qwen3-4B-Base on UltraFeedback, $\mu=1.0$, both principles applied, matched decoding):
\begin{itemize}
    \item \textbf{$N_1$-adaptive:} $K_t=\lceil N_1(q_t)\rceil$, a per-token support matched to the estimated plateau order;
    \item \textbf{Fixed top-$10$ (default):} $K_t=K_{\max}=10$, covering the low-order plateaus and the prominent higher-order regime around $k=10$ identified in Section~\ref{sec:interpreting_plateau_k};
    \item \textbf{Full vocabulary:} $S_t=\mathcal{V}$ and $\mathcal{A}_t=\mathcal{V}\setminus\{y_t\}$, covering all non-label tokens.
\end{itemize}
The three rows are ordered by how much of the base distribution they constrain, and they probe the two ends of the range separately.

Widening from fixed top-$10$ to the full vocabulary changes the average by only $0.22$ points ($58.79$ vs. $59.01$), far less than the degradation caused by dropping either design principle in Table~\ref{tab:lpsft_ablation_results}. This is what the probability weighting in Eq.~\eqref{eq:lp_loss} predicts: the additional tokens lie well outside the plateau, so the base model assigns them negligible weight and they can neither help nor hurt much. The costs differ sharply, however, since full-vocabulary preservation requires an in-memory frozen reference model throughout training (similar to ASFT~\citep{zhu2025anchored}), whereas fixed top-$10$ preservation relies entirely on a reusable offline cache (Table~\ref{tab:app_qwen25_72b_training_time}).

Narrowing behaves differently. The $N_1$-adaptive schedule ties the budget to the estimated plateau order at each position and is the weakest of the three ($57.74$). For confident predictions with $N_1(q_t) \le 2$, which constitute the majority of token positions, it sets $K_t \le 2$; whenever the supervised token $y_t$ falls within this small candidate set, the preservation set $\mathcal{A}_t = S_t \setminus \{y_t\}$ contains at most one token and the locally normalized KL loss vanishes identically. An entropy-indexed budget thus switches preservation off exactly where the model is confident. This is consistent with how we read Section~\ref{sec:interpreting_plateau_k}: plateau order indicates how many alternatives are clearly plausible, not how many are worth preserving, and the loss already down-weights implausible candidates on its own. Per-token adaptation therefore adds a moving part without adding selectivity, and we adopt the fixed $K_t=K_{\max}=10$ as the default in LP-SFT.

\section{Setup of the Collapse-Rate Diagnostic}
\label{app:collapse_diag}

We compute the collapse rate (Eq.~\eqref{eq:collapse_rate}) for the UltraFeedback-trained checkpoints in Table~\ref{tab:ultrafeedback_results}, each compared against its own base model. For each of the first 300 MATH-500 problems, the base model generates one solution (same prompt as in Appendix~\ref{app:training_eval_details}, temperature $0.6$, at most 1024 new tokens). The base model and each fine-tuned checkpoint then score the same problem--solution sequences, so $q_t$ and $p_t$ always share the same context. Special tokens are excluded, $\mathcal{A}_t$ follows Eq.~\eqref{eq:action_set}, and all probabilities are full-vocabulary without local normalization. The threshold $q_t(v)\ge0.05$ keeps only alternatives the base model finds plausible. The threshold $p_t(v)<10^{-3}$ marks a token that would be sampled fewer than once in 1{,}000 draws, far beyond our pass@$k$ budgets ($k\le32$). Table~\ref{tab:collapse_diag_counts} reports the number of positions and endorsed alternatives.

\begin{table}[h]
\centering
\small
\caption{Number of scored response positions and base-endorsed alternatives ($q_t(v)\ge0.05$, $v\in\mathcal{A}_t$) in the collapse-rate diagnostic, computed on 300 base-generated MATH-500 solutions per model family.}
\label{tab:collapse_diag_counts}
\begin{tabular}{lcc}
\toprule
\textbf{Base model} & \textbf{Positions} & \textbf{Endorsed alternatives} \\
\midrule
Qwen3-4B     & 87{,}151  & 40{,}227 \\
Qwen3-14B    & 135{,}365 & 39{,}001 \\
Llama-3.1-8B & 111{,}039 & 53{,}935 \\
\bottomrule
\end{tabular}
\end{table}

\end{document}